\documentclass[runningheads]{llncs}

\usepackage{eccv}

\usepackage[dvipsnames]{xcolor}

\usepackage{comment}

\usepackage{array,etoolbox}
\usepackage[utf8]{inputenc} 
\usepackage[T1]{fontenc}    

\usepackage{url}            
\usepackage{booktabs}       
\usepackage{amsfonts}       
\usepackage{nicefrac}       
\usepackage{microtype}      
\usepackage{xcolor}         

\usepackage{times}
\usepackage{epsfig}
\usepackage{graphicx}
\usepackage{amsmath}
\usepackage{amssymb}

\usepackage{bm} 
\usepackage{tabularx}
\usepackage{wrapfig}
\usepackage{bbm}
\usepackage{enumitem}
\usepackage{multicol}
\usepackage{booktabs}
\usepackage{svg}

\usepackage{comment}
\usepackage{array,etoolbox}
\usepackage{pifont} 

\usepackage{algorithm}
\usepackage{algpseudocode}
\usepackage{color, colortbl}

\usepackage[accsupp]{axessibility}  

\usepackage{eccvabbrv}

\preto\tabular{\setcounter{magicrownumbers}{0}}
\newcounter{magicrownumbers}

\DeclareMathOperator*{\argmax}{arg\,max}

\newcommand{\cmark}{\ding{51}}%
\newcommand{\xmark}{\ding{55}}%

\definecolor{Gray}{gray}{0.9}
\definecolor{LightCyan}{rgb}{0.88,1,1}

\usepackage[pagebackref,breaklinks,colorlinks,citecolor=eccvblue]{hyperref}

\usepackage{orcidlink}

\begin{document}

\title{Flexible Distribution Alignment: Towards Long-tailed Semi-supervised Learning with Proper Calibration}


\author{
Emanuel Sanchez Aimar\orcidlink{0000-0001-9874-737X},
Nathaniel Helgesen\orcidlink{0009-0003-4516-5685},\\
Yonghao Xu\orcidlink{0000-0002-6857-0152}, Marco Kuhlmann\orcidlink{0000-0002-2492-9872}, Michael Felsberg\orcidlink{0000-0002-6096-3648}\\
}


\authorrunning{E.~Sanchez Aimar et al.}

\institute{
Linköping University, Sweden\\
\{emanuel.sanchez.aimar,nathaniel.helgesen,yonghao.xu,marco.kuhlmann,michael.felsberg\}@liu.se
}

\maketitle

\begin{abstract}
Long-tailed semi-supervised learning (LTSSL) represents a practical scenario for semi-supervised applications, challenged by skewed labeled distributions that bias classifiers. This problem is often aggravated by discrepancies between labeled and unlabeled class distributions, leading to biased pseudo-labels, neglect of rare classes, and poorly calibrated probabilities. To address these issues, we introduce Flexible Distribution Alignment (FlexDA), a novel adaptive logit-adjusted loss framework designed to dynamically estimate and align predictions with the actual distribution of unlabeled data and achieve a balanced classifier by the end of training. FlexDA is further enhanced by a distillation-based consistency loss, promoting fair data usage across classes and effectively leveraging underconfident samples. This method, encapsulated in ADELLO (Align and Distill Everything All at Once), proves robust against label shift, significantly improves model calibration in LTSSL contexts, and surpasses previous state-of-of-art approaches across multiple benchmarks, including CIFAR100-LT, STL10-LT, and ImageNet127, addressing class imbalance challenges in semi-supervised learning. Our code is available at \url{https://github.com/emasa/ADELLO-LTSSL}.

\end{abstract}

\section{Introduction}
\begin{figure}[htbp]
\centering
\begin{subfigure}[tb]{0.24\textwidth}
    \includegraphics[width=\textwidth]{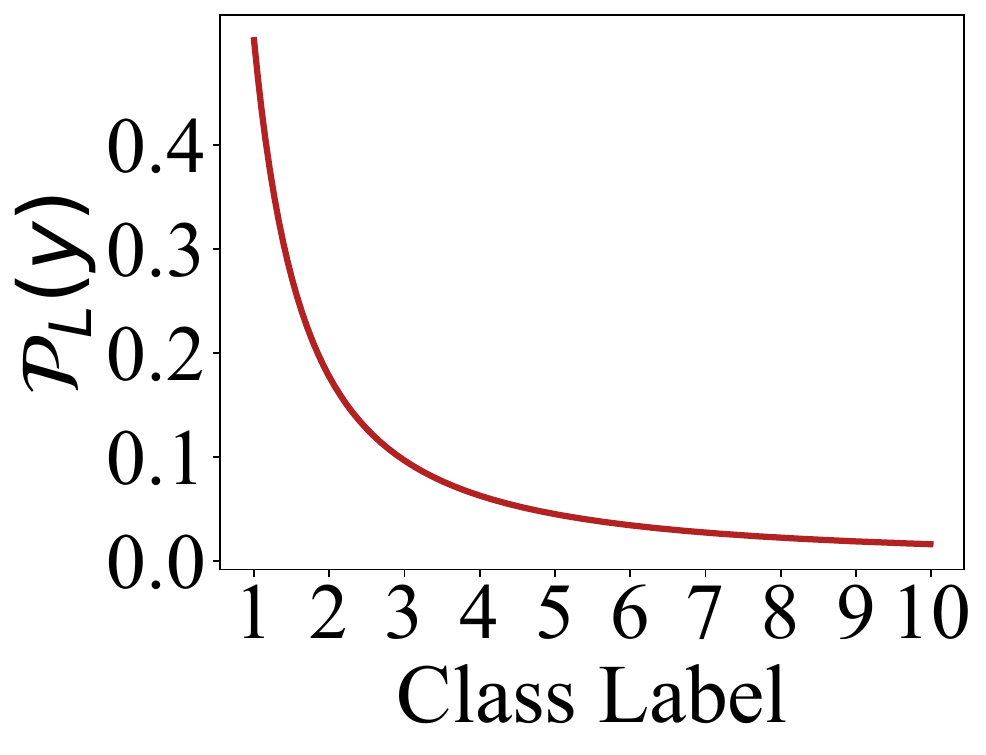}
    \caption{Labeled prior}   
    \label{fig:labeled_dist}
\end{subfigure}
\begin{subfigure}[tb]{0.24\textwidth}
    \includegraphics[width=\textwidth]{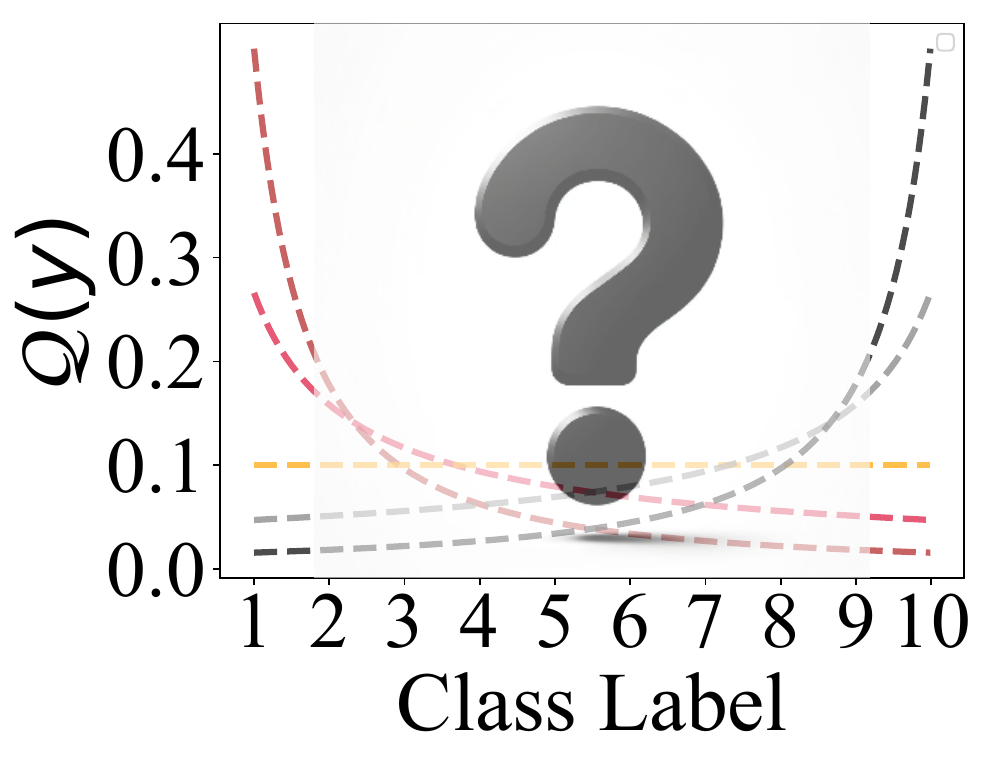}
    \caption{Unlabeled prior}
    \label{fig:unlabeled_dist}
\end{subfigure}
\begin{subfigure}[tb]{0.24\textwidth}
    \includegraphics[
    trim={0.2cm 0.3cm 0.3cm 0.2cm},clip,
    width=\linewidth]{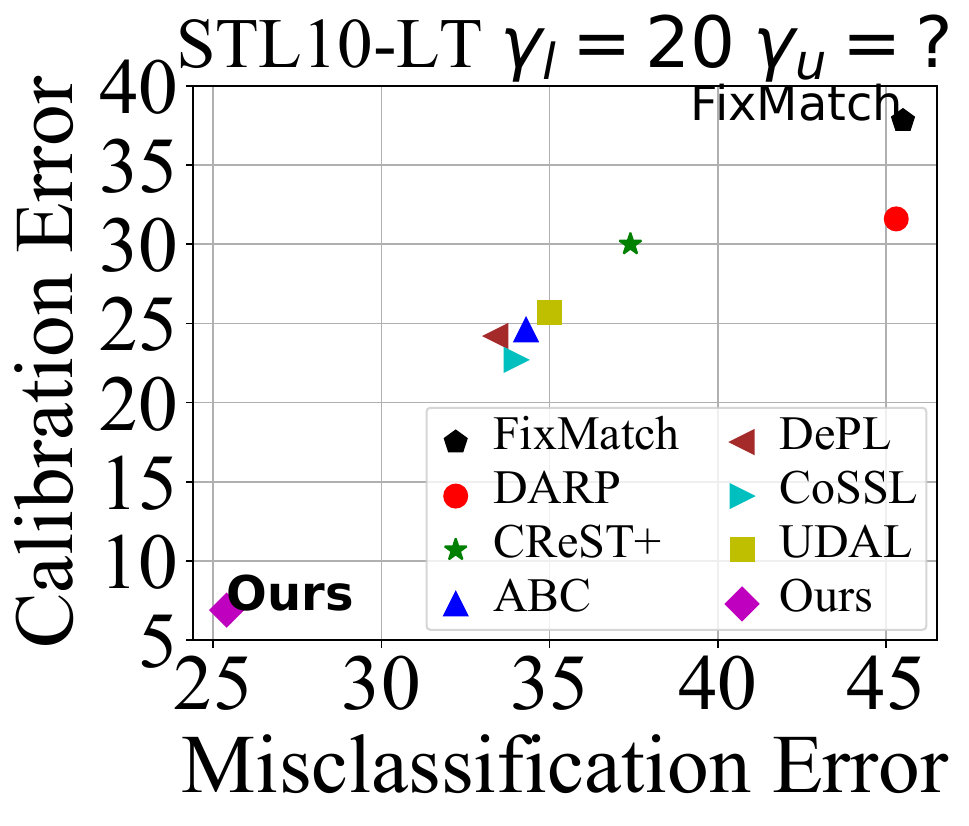}
    \caption{Generalization-Calibration}   
\label{fig:generalization_vs_calibration_top}
\end{subfigure}
\begin{subfigure}[tb]{0.24\textwidth}
    \includegraphics[
    trim={0.2cm 0.3cm 0.3cm 0.2cm},clip,
    width=\linewidth]{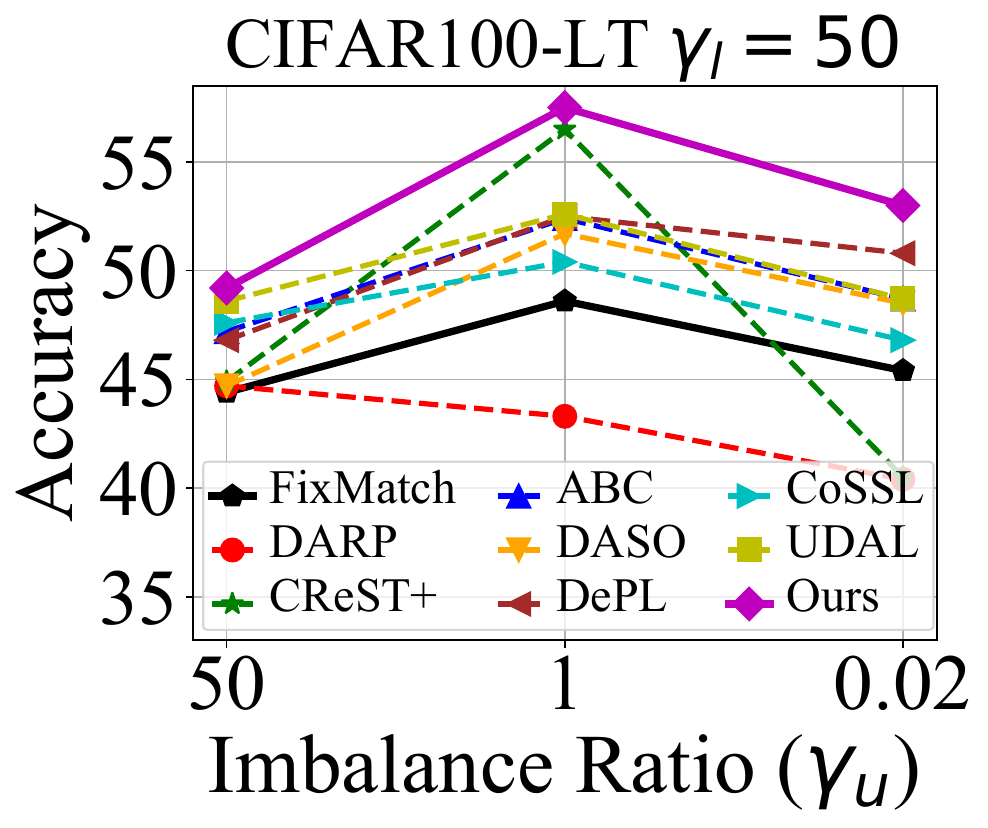}
    \caption{Robustness to label shift}
    \label{fig:label_shift_top}
\end{subfigure}
\caption{Long-tailed semi-supervised learning considers a challenging scenario, where a labeled dataset with a \textbf{skewed class distribution}, $\mathcal{P}_L(y)$, see \textbf{(a)}, can bias the model towards frequent classes. This challenge is exacerbated by the use of a larger unlabeled dataset with an \textbf{unknown class distribution}, $\mathcal{Q}(y)$, see \textbf{(b)}, risking the reinforcement of data biases, which in turn leads to \textbf{uncalibrated probabilities}. Our evaluation focuses on misclassification error (complement of accuracy) and expected calibration error to test generalization and calibration. Our approach shows consistent improvements in both respects, as shown in \textbf{(c)} and \textbf{(d)}.}
\label{fig:semi_supervised_dists}
\end{figure}

Solving computer vision tasks with limited labeled data is a challenging problem that has motivated the development of semi-supervised learning (SSL)~\cite{chapelle2006semi}. 
Training models on a mix of labeled and unlabeled data allows costly labeling to be circumvented, though unlabeled data has been shown to complicate training when the distribution of classes is highly imbalanced or follows a long-tailed distribution~\cite{powers1998applications}, as depicted in Fig.~\ref{fig:labeled_dist}. Notably, common SSL techniques, namely pseudo-labelling~\cite{lee2013pseudo} and high-confidence thresholding~\cite{sohn2020fixmatch}, can lead to imbalanced pseudo-label distributions even in balanced settings~\cite{wang2022debiasmatch}, producing classifiers that are biased towards head classes~\cite{wang2017tail,wang2022debiasmatch}.

Distribution alignment (DA)~\cite{berthelot2019remixmatch,lazarow2023udal} aims to mitigate these issues by aligning pseudo-label distributions with actual label priors in balanced~\cite{berthelot2019remixmatch,wang2022debiasmatch} and, more generally, long-tailed (LT) settings~\cite{wei2021crest,wang2022debiasmatch}. Despite recent progress in 
long-tailed semi-supervised learning(LTSSL)~\cite{kim2020darp,oh2021daso,fan2021cossl,lai2022saw}, most DA methods assume that labeled and unlabeled data follow the same distribution~\cite{berthelot2019remixmatch,wei2021crest,wang2022debiasmatch}, though in practical or low-label applications, the unlabeled class distribution is more likely to be unknown and distinct from the labeled distribution, as illustrated in Fig.~\ref{fig:unlabeled_dist}. This mismatch between labeled and unlabeled data, or \textit{label distribution shift}~\cite{hong2021lade}, can lead to \textit{low data utilization}, where models fail to effectively leverage the breadth of available data, especially samples from minority classes~\cite{kim2020darp,wei2021crest}. Therefore, incorporating the correct data prior is critical for improving model performance~\cite{kim2020darp, hong2021lade,zhang2022sade}. Previous DA approaches either require access to the ground truth prior or an approximation from balanced holdout data in advance for SSL training~\cite{berthelot2019remixmatch,kim2020darp,lazarow2023udal}, conditions that are often challenging to fulfill. Anchor distributions~\cite{wei2023acr,ma2023ltssl_experts} 
aim to sidestep these requirements, but challenges remain due to implicit assumptions about the class prior and the increased complexity of these approaches.

Model calibration, which guarantees that model predictions align with actual outcomes~\cite{guo2017calibration}, has gained recent attention for both enabling effective pseudo-labeling in SSL~\cite{loh2022calibrationssl} and facilitating distribution alignment in LT fully-supervised scenarios~\cite{menon2021longtail,xu2021bayias_calibrated,aimar2023balpoe_calibrated}. 
Beyond mitigating confirmation bias~\cite{loh2022calibrationssl}, calibration is crucial when using logit adjustment to address label shift~\cite{aimar2023balpoe_calibrated}. This highlights the importance of further investigating the influence of proper calibration on distribution alignment within LTSSL.

In this work, we tackle distribution misalignment in LTSSL, focusing on the shift between labeled and unlabeled class distributions. We present the following contributions:
\begin{itemize}
    \item \textbf{Flexible Distribution Alignment (FlexDA):} introduces novel supervised and consistency logit-adjusted losses to dynamically align the model with the class distribution of unlabeled data during training, boosting performance across diverse unlabeled class distributions. It integrates a progressive scheduler to adjust the target prior towards a balanced classifier gradually, crucial to accurate debiasing at inference.
    
    \item \textbf{Complementary Consistency Regularization:} augments FlexDA by distilling underconfident samples, optimizing the utilization of data often ignored due to low-confidence pseudo-labels. 
    
    \item \textbf{Study of Model Calibration in LTSSL:} We delve into the relationship between calibration and generalization in LTSSL, corroborating that improved calibration is highly correlated with enhanced model generalization across various datasets and different degrees of label shift.
\end{itemize}
The effectiveness of our methodology is showcased in settings with both controlled and unknown label shifts, such as CIFAR100-LT and STL10-LT, where it excels in LTSSL contexts, see Fig.~\ref{fig:label_shift_top} and Fig.~\ref{fig:generalization_vs_calibration_top}. Under matching distributions, it surpasses state-of-the-art (SOTA) approaches in the large-scale dataset ImageNet127. Finally, ADELLO presents significant advances in calibration compared to previous LTSSL approaches, see Fig.~\ref{fig:generalization_vs_calibration_top}. These contributions form the foundation of \textit{Align and Distill Everything All at Once} (ADELLO), outlined in Fig.~\ref{fig:adello-diagram}, a versatile framework adept at handling distribution shifts, efficiently overcoming LTSSL challenges, and enhancing model calibration.

\section{Related Work}
\label{sec:related_work}
\textbf{Semi-supervised learning.}
Semi-supervised learning is a mature field, with significant advancements made in recent decades \cite{scudder1965firstpl,mclachlan1975reclass,grandvalet2005minentropy,chapelle2006semi,lee2013pseudo,sohn2020fixmatch}. Successful techniques such as pseudo-labeling \cite{lee2013pseudo,arazo2020plcb}, consistency-regularization \cite{tarvainen2017meanteachers, miyato2018vat}, and their combinations \cite{berthelot2019mixmatch,berthelot2019remixmatch, kuo2020featmatch,arazo2020plcb} have contributed to this progress. Encouraging consistency between weakly-perturbed data views has shown improvements \cite{sajjadi2016firstconsistregular,laine2016pimodel}, and further progress has been made with the combination of weak and strong data augmentation \cite{xie2019uda,sohn2020fixmatch}.

Confidence-based methods combine high-confidence thresholding and pseudo-labeling (hard \cite{sohn2020fixmatch} or temperature-scaled \cite{xie2019uda}) to minimize confirmation bias \cite{kraus2010rim}. However, a fixed threshold can limit the utilization of unlabeled samples \cite{Lucas2022barely_supervised,anonymous2023softmatch}, especially in low-labeled regimes. 
In response, later approaches introduced progressive thresholds \cite{xu2021dash} and class-specific thresholds \cite{zhang2021flexmatch}. Recently, SoftMatch \cite{anonymous2023softmatch} integrates adaptive thresholding and confidence-based weighting for better sample utilization. Contrasting with these methods is the \textit{knowledge distillation} (KD) approach, characterized by the use of softened output logits to distill information about class similarities \cite{hinton2015distilling}. In contexts with minimal labeled data, self-supervised prototypes, acquired via online deep-clustering~\cite{caron2020online_clustering}, are distilled to exploit predictions below an adaptive threshold~\cite{Lucas2022barely_supervised}. Diverging from~\cite{Lucas2022barely_supervised,anonymous2023softmatch}, our method uniquely distills soft pseudo-labels that fall below a threshold, this process being steered by a distribution alignment loss. 

To mitigate the bias induced by imbalanced pseudo-labeled distributions,  some techniques aim to align the model distribution with the labeled prior, often uniform, based on maximum mean-entropy regularization~\cite{arazo2020plcb, kraus2010rim,wang2022freematch}. Additionally, other distribution alignment approaches involve correcting the prior directly on pseudo-labels \cite{wang2022debiasmatch, berthelot2019remixmatch, berthelot2021adamatch} and using margin-based losses to debias the classifier~\cite{wang2022debiasmatch}.

\textbf{Long-tailed recognition.} 
In real-world datasets, LT distributions are common, where a few classes dominate with numerous examples, and most have significantly fewer~\cite{van2017devil, Zhu2014CVPR}. Addressing this imbalance, methods based on data resampling~\cite{chawla2002smote,kubat1997addressing,wallace2011class, hyun2020class}, loss reweighting~\cite{lin2017focal, huang2016learning_inverse_frecuency,cui2019class_effective_number, tan2020equalization}, and margin modifications~\cite{xie1989logit, morik1999combining,cao2019ldam} have been developed. Theoretically-grounded logit adjustments (LA) mitigate the LT bias~\cite{menon2021longtail}, yielding balanced~\cite{ren2020balanced, menon2021longtail} and well-calibrated classifiers \cite{xu2021prior_calibrated, aimar2023balpoe_calibrated}. These adjustments can be applied during the optimization~\cite{ren2020balanced, menon2021longtail} or as post-hoc bias correction~\cite{menon2021longtail, hong2021lade}. Additionally, expert-based models are specialized in handling either single or multiple target distributions~\cite{li2022nested_experts, zhang2022sade, aimar2023balpoe_calibrated} and have the capability to adjust the test target prior by leveraging unlabeled data transductively~\cite{zhang2022sade}. Lastly, KD~\cite{hinton2015distilling} can also be applied for transferring knowledge from head to tail classes~\cite{xiang2020distilling_lt_experts, He2021DIVE_virtual_examples, li2022nested_experts}.

\textbf{Long-tailed semi-supervised learning.}
SSL research has recently focused on long-tailed scenarios by relaxing the uniform assumption~\cite{kim2020darp,wei2021crest}. DARP~\cite{kim2020darp} refines pseudo-labels via convex optimization, while CReST~\cite{wei2021crest} reduces class imbalance by expanding the labeled dataset with unlabeled data across multiple generations. Class-dependent approaches weight losses based on class difficulty~\cite{lai2022saw} and varying pseudo-label thresholds based on relative class-frequencies~\cite{guo2022class}. Several approaches introduce auxiliary balanced classifiers~\cite{lee2021abc,oh2021daso,fan2021cossl,wei2023acr} or feature regularization~\cite{fan2021cossl} to address LT issues.

CReST+~\cite{wei2021crest} introduces a schedule for progressively aligning the distribution of pseudo-labels, transitioning from long-tailed to more balanced distributions, a strategy later incorporated as part of the training loss~\cite{lazarow2023udal}. However, most DA approaches assume that labeled and unlabeled data share similar marginal distributions, which may not always hold, and often requires additional supervised pre-training to estimate distribution mismatch~\cite{kim2020darp, lazarow2023udal}. Diverging from this, some studies have adopted re-weighting strategies across fixed anchor distributions~\cite{wei2023acr,ma2023ltssl_experts}. However, these methods might inadvertently rely on privileged information, as they constrain the adjustment to the family of label distributions typically used in evaluation benchmarks. In contrast, our approach utilizes distribution alignment losses, dynamically adjusting to the class distribution of unlabeled data. This reduces label bias and aims for an unbiased, balanced classifier during inference, without being confined to a specific family of prior distributions. Additionally, our method leverages all data samples for regularization, enhancing both prior estimation and distribution alignment, further detailed in Sections~\ref{align} and~\ref{distill}.

\textbf{Confidence calibration.} Over-parameterized networks tend to yield uncalibrated and overly confident predictions, particularly in the presence of out-of-distribution data~\cite{guo2017calibration,li2020wrongly_overconfident}, a problem exacerbated by class imbalances~\cite{zhong2021mislas_mixup}. While Mixup~\cite{zhang2017mixup} and its variants~\cite{verma2019manifold,yun2019cutmix} improve calibration in fully-supervised settings and adaptations for long-tailed contexts exist~\cite{xu2021bayias_calibrated,park2022cmo_cutmix_lt}, integrating these approaches with threshold-based methods, such as FixMatch~\cite{sohn2020fixmatch}, remains challenging. Nevertheless, the crucial link between calibration and model performance in balanced SSL setups~\cite{loh2022calibrationssl} underscores the importance of calibration. Finally, fully supervised LT distribution alignment methods critically hinge on well-calibrated probabilities~\cite{menon2021longtail,hong2021lade,aimar2023balpoe_calibrated}, underscoring the urgency of addressing calibration within LTSSL frameworks.

\section{Preliminaries}
\label{sec:background}
\textbf{Problem formulation.}
In long-tailed semi-supervised learning for classification tasks, we address a scenario with a limited labeled dataset $D_L = (X_L, Y_L)$ and a larger unlabeled dataset $D_U = (X_U, \cdot)$. The labeled dataset comprises $N$ samples $(x_i, y_i)$, where $y_i \in \{1, \dots, K\}$ denotes the class labels and $K$ is the total number of classes. The unlabeled dataset contains $M$ $(u_i, \cdot)$ samples, following an unknown class distribution $\mathcal{Q}(y)$. Classes are sorted by descending order of labeled sample size, i.e., $N_1 \ge N_2 \ge ... \ge N_K$, where $N_k$ represents the number of labeled samples for class $k$. The imbalance ratio for the labeled set, $\gamma_l$, is defined as $N_1/N_K$. Similarly, the (unknown) imbalance ratio for the unlabeled set, $\gamma_u$, is $M_1/M_K$, where $M_k$ is the count of unlabeled samples in class $k$. The objective is to train a classifier that minimizes the \textit{balanced error rate} (BER) \cite{chan1998balancederror,brodersen2010balancederror} on the test distribution, thus ensuring fair treatment of minority classes.

For labeled data, the standard supervised loss \cite{sohn2020fixmatch,xie2019uda}, denoted by $\mathcal{L}_s$, is the average cross-entropy, $ \mathcal{H}(\cdot,\cdot)$, between the true labels $y$ and the model predictions $p(y|x)$, 
\begin{equation}
\mathcal{L}_{s} = \frac{1}{B} \sum \nolimits^{B} _{b=1} \mathcal{H}(y_b, p(y|\omega(x_b))),
\label{eq:base_loss_sup_lab}
\end{equation}
where $p(y|x) = \sigma( f(x) )$ denotes the prediction produced by a neural network $f$, normalized by the \textit{softmax} function, $\sigma(\cdot)$; $\omega(\cdot)$ denotes a weak data augmentation procedure and $B$ denotes the batch size. 

For unlabeled data, we are interested in threshold-based consistency regularization approaches \cite{xie2019uda,sohn2020fixmatch,zhang2021flexmatch,wang2022freematch}. Following \cite{xie2019uda,sohn2020fixmatch}, we employ weak and strong data augmentations, denoted by $\omega(\cdot)$ and $\Omega(\cdot)$, respectively. The unsupervised loss is defined as
\begin{equation}
\mathcal{L}_{u} = \frac{1}{\mu B} \sum \nolimits^{\mu B}_{b=1} \mathcal{M}(u_b) \cdot \mathcal{H}(\hat{y}_b, p(y|\Omega(u_b))),
\label{eq:base_loss_unl}
\end{equation}
where $\mathcal{M}(u_b) = \mathbbm{1}(\max(p(y|\omega(u_b)) \ge \tau)$ denotes a sample mask relative to a threshold~$\tau$, $\hat{y}_b = \argmax p(y|\omega(u_b))$ is a one-hot pseudo-label \cite{sohn2020fixmatch}. $\mu$ determines the relative sizes of labeled and unlabeled data in a batch.

\textbf{Distribution alignment in SSL.} Table \ref{tab:DA_comparison} contrasts various methods of distribution alignment tailored for SSL in scenarios with class imbalance. Originally introduced for the balanced setting, ReMixMatch \cite{berthelot2019remixmatch} aligns the marginal distribution of predictions on unlabeled data, estimated by an exponential moving average (EMA), $ \hat{Q}(y) \approx \mathbb{E}_{u \sim X_U}[p(y|\omega(u))]$, with the labeled prior distribution $ \mathcal{P}_L(y) $, via pseudo-label adjustment (PL). CReST+ \cite{wei2021crest} adopts a more flexible multi-generational training approach for LT scenarios, controlling the rate and extent of PL debiasing to preserve the model precision and recall for unlabeled data. However, using only pseudo-label correction in long-tailed data scenarios can bias the classifier towards head classes, even with correct pseudo-labels, as noted in fully-supervised settings \cite{kang2019decouple,menon2021longtail}.

\begin{wraptable}{r}{0.48\linewidth} 
\vspace{-30pt}
\centering
\caption{\textbf{Comparative overview of distribution alignment methods under class imbalance}. ``PL'' denotes pseudo-label adjustment; ``S loss'' refers to supervised logit-adjusted loss; ``U loss'' indicates unsupervised logit-adjusted loss; ``LT'' denotes long-tailed labeled prior; ``Progressive'' describes gradual smoothing of the target prior throughout training. 
}
\label{tab:DA_comparison}
\resizebox{0.48\textwidth}{!}{
\begin{tabular}{@{}lllc@{}}
\toprule
Method & Strategy & Target Prior & Progressive \\ \midrule
ReMixMatch \cite{berthelot2019remixmatch} & PL & LT & \xmark \\
CReST+ \cite{wei2021crest} & PL & LT & \cmark \\
LA \cite{menon2021longtail,ren2020balanced} & S loss & Uniform & \xmark \\
DebiasPL \cite{wang2022debiasmatch} & PL \& U loss & Uniform & \xmark \\
UDAL \cite{lazarow2023udal} & S \& U losses & LT & \cmark \\ 
ADELLO (ours) & S \& U losses & \textbf{Adaptive} & \cmark
\\ \bottomrule
\end{tabular}
}
\vspace{-20pt}
\end{wraptable}

In the fully-supervised case, logit-adjusted losses \cite{menon2021longtail,ren2020balanced} correct the long-tailed label bias by aligning the distribution towards a uniform prior. 
DebiasPL \cite{wang2022debiasmatch} combines pseudo-label generation with a margin-based unsupervised loss, controlled by a static hyper-parameter. 
UDAL \cite{lazarow2023udal}, inspired by CReST+, progressively adjusts its losses to target a smooth long-tailed prior \( \mathcal{P}_{\alpha_t}(y) \), achieving a more effective alignment.
These methods typically presuppose that the unlabeled marginal distribution \( \mathcal{Q}(y) \) is similar to the labeled distribution \( \mathcal{P}_L(y) \). In situations where the target distribution \( \mathcal{Q}(y) \) is unknown, previous research \cite{kim2020darp, lazarow2023udal} has proposed modifying the consistency loss to include a predefined target distribution. However, this approach can be challenging, particularly in the absence of prior knowledge about the true distribution or when a balanced, labeled hold-out dataset for estimating the prior \cite{lipton2018detecting} is not available.

Aiming to resolve these challenges, ADELLO aligns the target distribution to the unknown unlabeled data and fosters robust consistency by distilling even low-confident predictions simultaneously to counter the class imbalance.

\section{ADELLO framework}
\label{sec:method}

\begin{figure*}[t]
    \centering
    \includegraphics[trim={0.5cm 4cm 0.5cm 2.5cm},clip,width=0.8\linewidth]{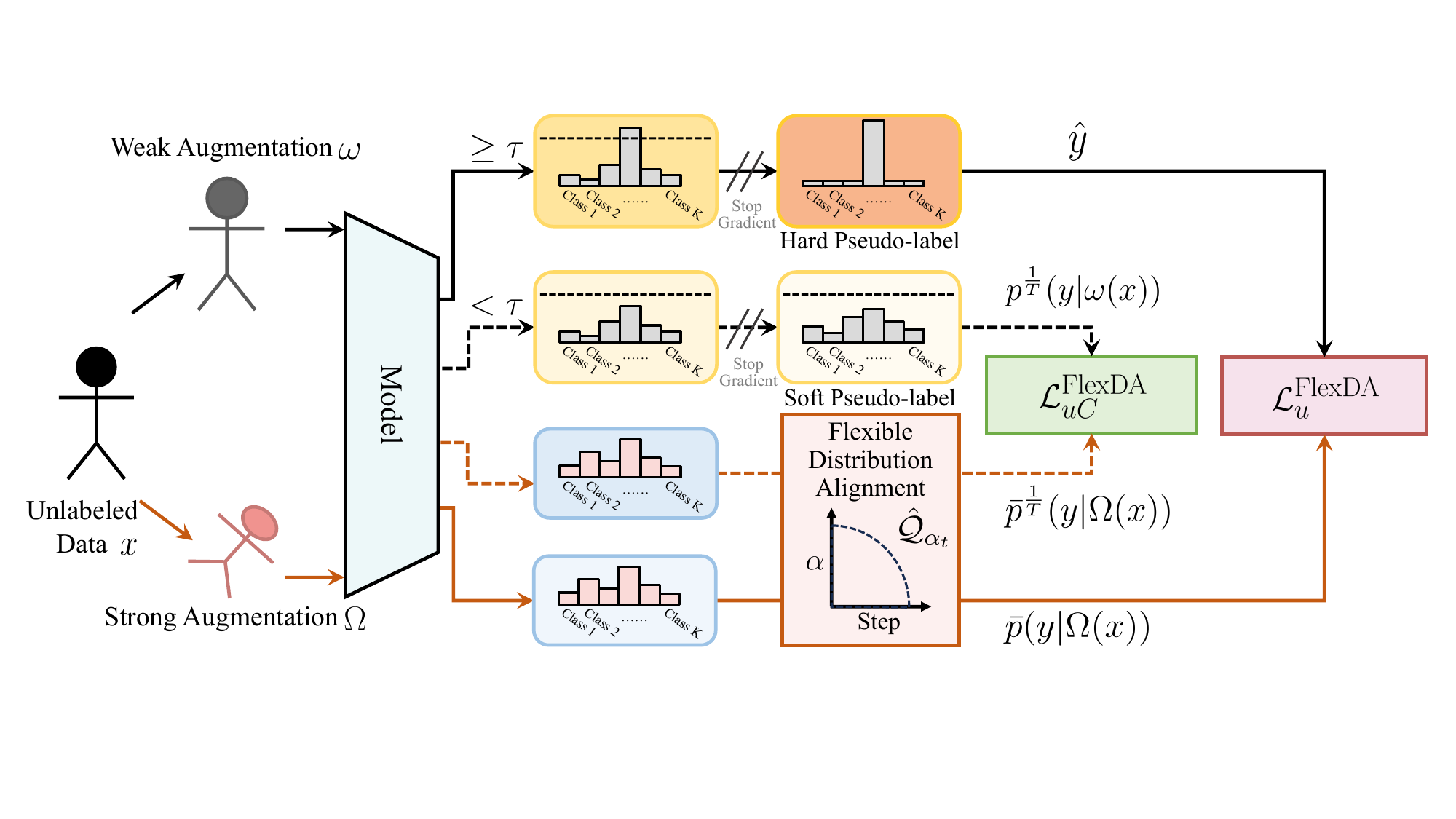}
    \caption{\textbf{Method overview}: Our \textit{flexible distribution alignment} (FlexDA) aligns the classifier with the correct prior, dynamically estimated from unlabeled data. This approach extends FixMatch with a bias-adjusted supervised loss ((\ref{eq:flexda_s}), Sec.~\ref{align}) and 
    a bias-adjusted consistency loss ((\ref{eq:flexda_u}), Sec.~\ref{align}) to debias high-confidence \textit{hard pseudo-labels}. We also introduce a bias-adjusted \textit{complementary consistency} loss to learn from low-confidence \textit{soft pseudo-labels} (Sec.~\ref{distill}). A progressive scheduler steadily smooths the target prior, $ \mathcal{\hat{Q}}_{\alpha_t} $, leading to a balanced classifier by the conclusion of training.}
    \label{fig:adello-diagram}
\end{figure*}

As illustrated in Fig.~\ref{fig:adello-diagram}, the key idea of ADELLO is to conduct progressive distribution alignment on unlabeled data via consistency regularization using high-confidence and low-confidence pseudo-labels. We propose the following objective,
\begin{equation}
\mathcal{L} = \mathcal{L}_s^{\text{FlexDA}} +\mathcal{L}_{u}^{\text{FlexDA}} + \mathcal{L}_{uC}^{\text{FlexDA}}\,,
\end{equation}
where $\mathcal{L}_s^{\text{FlexDA}}$ denotes the supervised loss (\ref{eq:flexda_s}), $\mathcal{L}_{u}^{\text{FlexDA}}$
denotes the unsupervised consistency loss (\ref{eq:flexda_u}), 
and $\mathcal{L}_{uC}^{\text{FlexDA}}$ denotes the complementary consistency loss (Sec. \ref{distill}) within FlexDA. This framework facilitates model alignment with an adaptive target distribution and effective use of labeled and unlabeled data, ensuring robust model training through comprehensive data utilization. We will now provide a detailed description of the losses introduced in ADELLO.

\subsection{Flexible Distribution Alignment} \label{align}

In the training of modern SSL frameworks, we typically observe two learning phases: 1) a supervised phase where the model is trained on a small labeled dataset using weak data augmentation, and 2) a phase where pseudo-labeling and consistency regularization are employed to learn from strongly-augmented unlabeled data. 

Let us assume that the first phase yields a scorer function $ g_L(x)$ that perfectly fits the labeled distribution, represented as $ g_L(x) \propto \mathcal{P}_L(y | x) $. 
In the second phase, we want to find the classifier that maximizes the number of correct pseudo-labels. The Bayes-optimal classifier emerges as a solution:
$\hat{y} = \argmax_y \mathcal{Q}(y | x) = \argmax_y \mathcal{Q}(x | y) \cdot \mathcal{Q}(y)$.
Under \textit{label shift} \cite{hong2021lade}, where priors might differ, i.e. $ \mathcal{P}_L(y) \neq \mathcal{Q}(y) $, yet the likelihood remains the same, i.e. $ \mathcal{P}_L(x | y) = \mathcal{Q}(x | y) $, we can define an adjusted scorer for pseudo-labeling: $ g_U(x) = g_L(x) \cdot \frac{\mathcal{Q}(y)}{\mathcal{P}_L(y)} $, and show that $ g_U(x) \propto \mathcal{Q}(y | x) $. Furthermore, this indicates that $ g_U(x) $ is the best scorer for the unlabeled data in terms of Bayes optimality. 

The definition of $ g_U(x) $ suggests that, in practice, obtaining a good classifier involves two challenges: neutralizing bias from the skewed labeled data and adjusting for the marginal distribution $ \mathcal{Q}(y) $. However, for inference, our goal is to achieve a \textit{balanced classifier} that treats all classes equally. This classifier should minimize the balanced error, which is independent of training or test priors:
$\hat{y} = \argmax_y \mathcal{P}_{\text{bal}}(y | x) = \argmax_y \mathcal{P}_{\text{bal}}(x | y) \cdot \mathcal{P}_{\text{bal}}(y) $. Here, $ \mathcal{P}_{\text{bal}}(y) = \frac{1}{K} $ denotes the uniform prior. A balanced scorer can thus be defined as $ g_B(x) = g_U(x) \cdot \frac{\mathcal{P}_{\text{bal}}(y)}{\mathcal{Q}(y)} \propto \mathcal{P}_{\text{bal}}(y| x) $, aligning with the Bayes-optimal rule for minimizing the BER \cite{menon2021longtail}.

Reconciling training demands with inference realities presents a challenge, as $ \mathcal{Q}(y) $ is often unknown. Even under consistent scenarios, $ \mathcal{P}_L(y) $ may be a biased estimate of the correct distribution, particularly with limited labeled samples. To address these challenges, we introduce the \textbf{Flex}ible \textbf{D}istribution \textbf{A}lignment (FlexDA) approach. FlexDA dynamically adapts to the characteristics of unlabeled data by aligning the model with a target distribution based on the EMA of the pseudo-labels during optimization. It employs a smoothed target prior, $ \mathcal{\hat{Q}}_{\alpha_t}(y) = \frac{\mathcal{\hat{Q}}(y)^{\alpha_t}}{\sum_j \mathcal{\hat{Q}}(j)^{\alpha_t}} $, with a time-updated decay factor $ \alpha_t $. Our approach progressively smooths the target prior, starting from the unlabeled prior $ \mathcal{\hat{Q}}_{1}(y) = \mathcal{\hat{Q}}(y) $, and gradually transitioning to a (near) balanced prior $ \mathcal{\hat{Q}}_{0}(y) = \frac{1}{K} $ by the end of the training process. 

In the FlexDA approach, our proposed logit-adjusted supervised loss is defined as
\begin{equation}
\label{eq:flexda_s}
\mathcal{L}_{s}^{\text{FlexDA}} = \frac{1}{B} \sum \nolimits^{B}_{b=1} \mathcal{H}(y_b, \sigma(f(\omega(x_b)) + \log \frac{\mathcal{P}_L}{\hat{\mathcal{Q}}_{\alpha_t}} )), 
\end{equation}
while our unsupervised consistency loss is defined as
\begin{equation}
\label{eq:flexda_u}
\mathcal{L}_{u}^{\text{FlexDA}} = \frac{1}{\mu B} \sum \nolimits^{\mu B}_{b=1} \mathcal{M}(u_b) \cdot \mathcal{H}(\hat{y}_b, \sigma( f(\Omega(u_b)) + \log \frac{\hat{Q}}{\hat{\mathcal{Q}}_{\alpha_t}} )), 
\end{equation}
where $ \alpha_t = 1.0 - (1.0 - \alpha_{\text{min}} ) (\frac{t}{t_\text{total}})^d $ defines a schedule. Here, $ t $ is the current training step, $ t_\text{total} $ the total number of training steps, and $ d $ and $ \alpha_{\text{min}} $ are hyper-parameters controlling the speed of the debiasing schedule and the minimum bias allowed, respectively.

\begin{figure*}[htbp]
\centering
\begin{subfigure}{.24\textwidth}
  \centering
  \includegraphics[width=\linewidth]{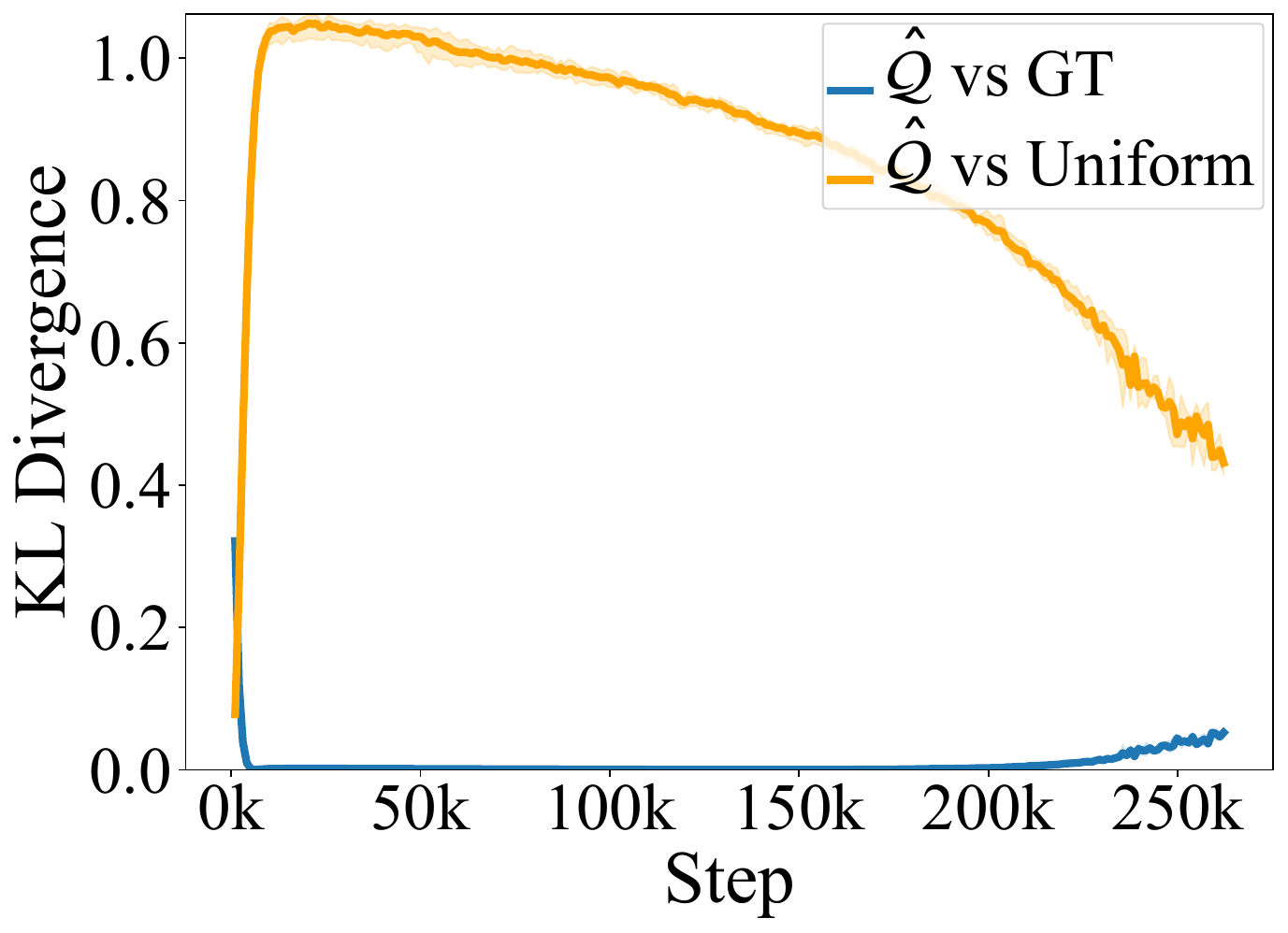}
  \caption{KL-div for \textit{forward} case}
  \label{fig:forward}
\end{subfigure}%
\begin{subfigure}{.24\textwidth}
  \centering
  \includegraphics[width=\linewidth]{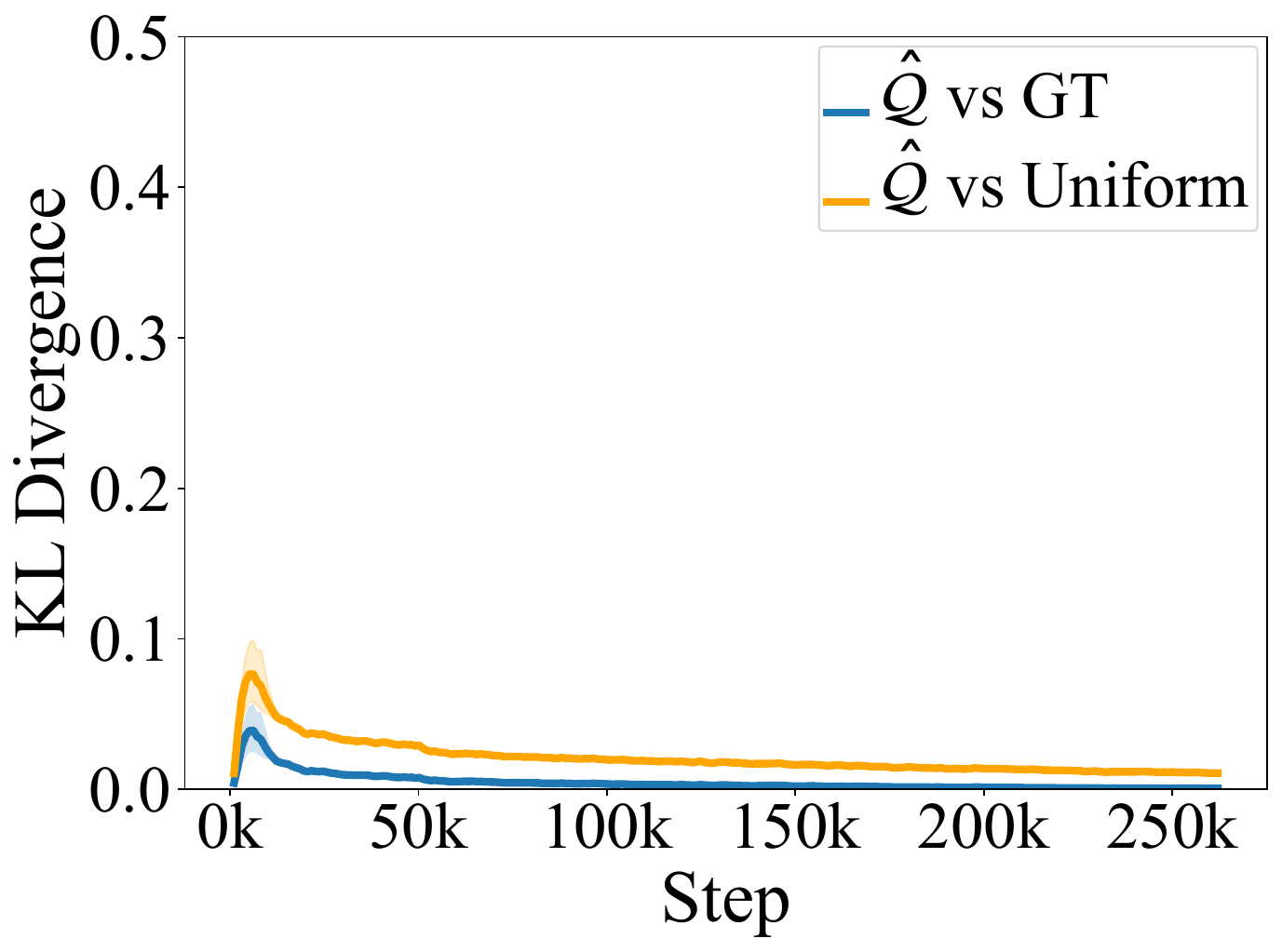}
  \caption{KL-div for \textit{balanced} case}
  \label{fig:balanced}
\end{subfigure}%
\begin{subfigure}{.24\textwidth}
  \centering
  \includegraphics[width=\linewidth]{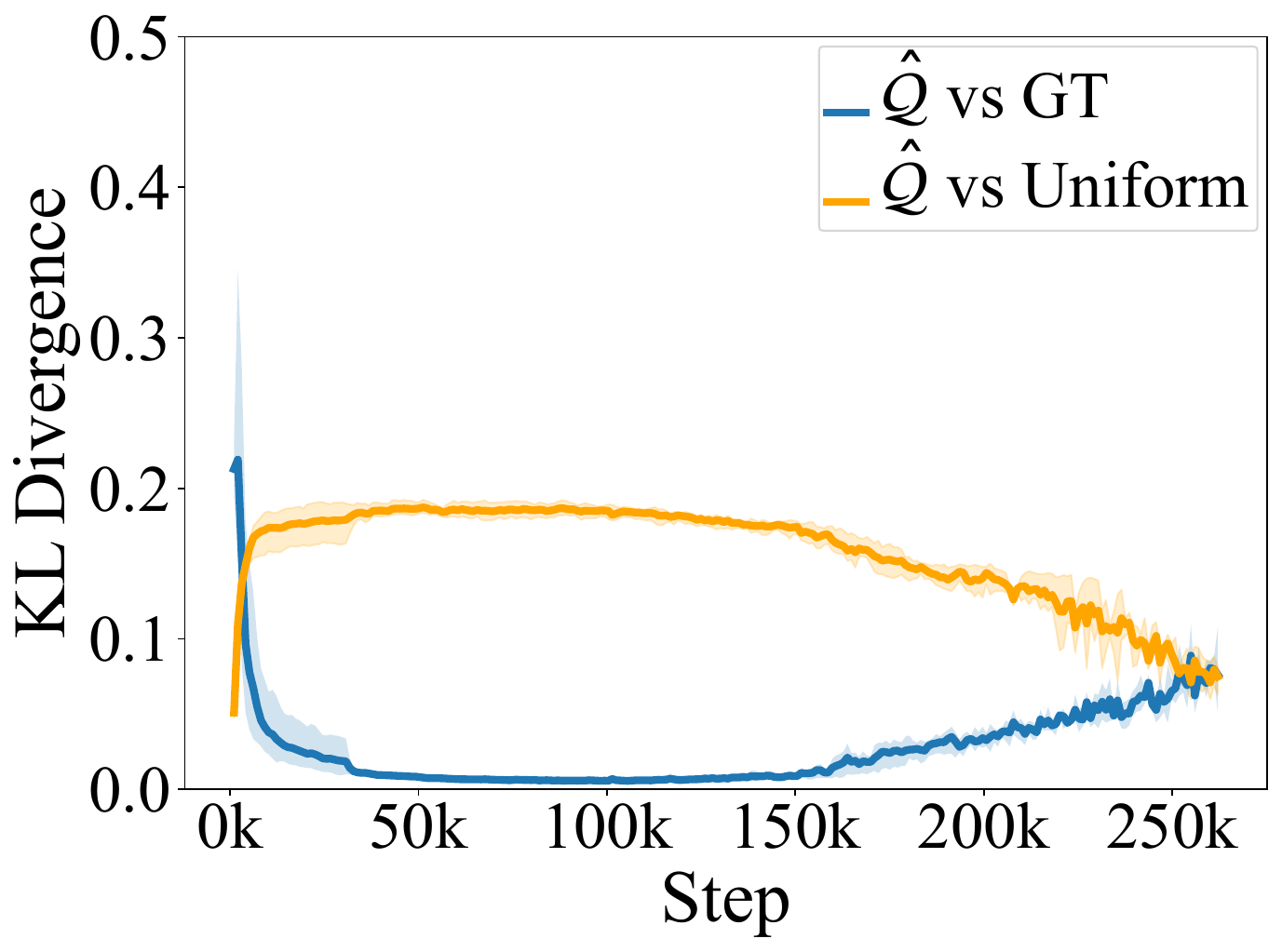}
  \caption{KL-div for \textit{reversed} case}
  \label{fig:reverse}
\end{subfigure}
\begin{subfigure}{.24\textwidth}
  \centering
  \includegraphics[width=\linewidth]{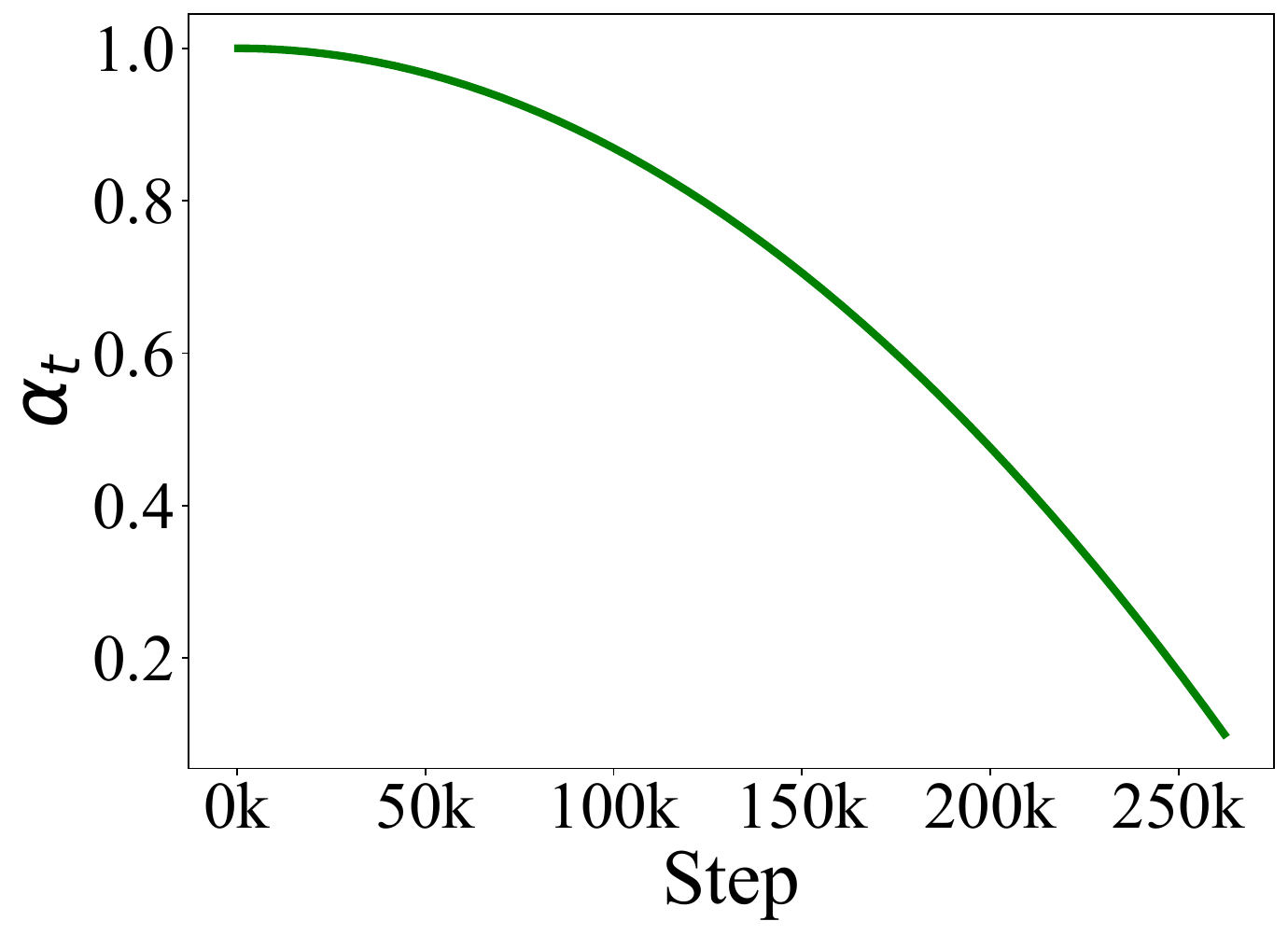}
  \caption{$\alpha_t$ ($d=2$, $\alpha_{\text{min}}=0.1$)}
  \label{fig:alpha_as_funtion_of_time}  
\end{subfigure}
\caption{\textbf{Prior estimation under label shift.} A comparison of KL divergence shows 1) a small difference between the estimated prior, \(\hat{\mathcal{Q}}\), and the ground-truth prior, \(\mathcal{Q}\), during most of the training (\textcolor{blue}{blue curve}), and 2) a larger disparity between \(\hat{\mathcal{Q}}\) and the uniform prior, \(\mathcal{P}_{\text{bal}}\), (\textcolor{orange}{orange curve}). The progression of a quadratic scheduler ($d = 2$) is shown in (d) (\textcolor{ForestGreen}{green curve}). Label shift settings: (a) forward, (b) balanced, and (c) reversed long-tailed, computed for CIFAR10-LT100.}
\label{fig:kl_divergence_priors}
\end{figure*}

\textbf{Statistical perspective.} For simplicity, let us assume $\alpha_{\text{min}} = 0$, i.e. a uniform target prior by the end of training. When $ \alpha_t \to 1$, our supervised loss compensates for the distribution shift between labeled and unlabeled data. While the unadjusted scorer might effectively classify labeled data, as in $ \sigma(f(x) + \log \frac{\mathcal{P}_L}{\hat{\mathcal{Q}}_{\alpha_t}} ) \overset{{\alpha_t \to 1}}{\approx} g_L(x) $, the model (adjusted scorer) aims to emulate the Bayes-optimal classifier for unlabeled data, i.e., $ \sigma(f(x)) {\approx} g_L(x) \cdot \frac{\hat{\mathcal{Q}}_{\alpha_t}}{\mathcal{P}_L} \overset{{\alpha_t \to 1}}{\approx} g_U(x) $. As Fig.~\ref{fig:kl_divergence_priors} shows, the estimated model prior $\hat{\mathcal{Q}}$ aligns closely with the ground-truth unlabeled prior $\mathcal{Q}$ from the beginning of training under various levels of label shift, as shown by the 
KL divergence between these distributions. 

Throughout the training, both supervised and unsupervised losses progressively reduce the bias induced by the unlabeled data, modulated by $ \alpha_t $ (refer to Fig.~\ref{fig:alpha_as_funtion_of_time}). By the end of the training, as $\alpha_t \to 0$ and $\hat{\mathcal{Q}}_{\alpha_t} \to \mathcal{P}_{\text{bal}}(y)$, FlexDA steers the model towards a (more) balanced distribution, leading to $ \sigma(f(x)) {\approx} g_U(x) \cdot \frac{\hat{\mathcal{Q}}_{\alpha_t}}{\hat{\mathcal{Q}}(y)} \overset{{\alpha_t \to 0}}{\approx} g_B(x) $. This dual loss structure effectively counters the bias introduced by labeled and unlabeled data samplers and adapts to the dynamic target prior. It provides a holistic approach for training semi-supervised models in the presence of class imbalance.

\subsection{Complementary Consistency Regularization} \label{distill}
The lack of ground-truth labels for tail classes complicates the generation of accurate pseudo-labels early in training, leading to a reduced supervisory signal when high-confidence thresholds are applied \cite{Lucas2022barely_supervised}. Furthermore, (progressive) distribution alignment can decrease the confidence of pseudo-labels as training progresses \cite{wei2021crest}, further weakening the supervisory signal. To utilize all available unlabeled data, we enhance the conventional consistency objective (\ref{eq:base_loss_unl}) with a complementary consistency regularization (CCR) technique. We implement a masked-distillation loss that makes use of soft pseudo-labels that fall below the confidence threshold:
\begin{equation}
\label{eq:complementary_consistency}
\mathcal{L}_{uC} = \frac{1}{\mu B} \sum \nolimits^{\mu B}_{b=1} \mathcal{M}^C(u_b) \cdot 
\mathcal{H}(p^{\frac{1}{T}}(y|\omega(u_b)), p^{\frac{1}{T}}(y|\Omega(u_b))),
\end{equation}
where 
$\mathcal{M}^C(u_b) = 1 - \mathcal{M}(u_b)$
denotes the complementary mask, $T$ is the temperature scaling factor, and $p^{\frac{1}{T}}(y|x) = \sigma ( \frac{1}{T} f(x) )$ represents the temperature-scaled predictions.

When combining our distribution alignment with complementary consistency, we obtain 
$\mathcal{L}_{uC}^{\text{FlexDA}}$ which is defined the same as $\mathcal{L}_{uC}$ but replacing $p^{\frac{1}{T}}(y|\Omega(u_b)))$ with $\bar{p}^{\frac{1}{T}}(y|\Omega(u_b)) = \sigma ( \frac{1}{T} (f(\Omega(u_b)) + \log \frac{\hat{Q}}{\hat{\mathcal{Q}}_{\alpha_t}}))$, where $\bar{p}^{\frac{1}{T}}$ denotes temperature-scale unadjusted predictions.
For simplicity, we denote $\bar{p}(y|x) = \bar{p}^{\frac{1}{T}}(y|x)$ when $T = 1$.

Contrasting with recent methods that sharpen 
(hence low temperature) low-confidence, less reliable pseudo-labels \cite{berthelot2019remixmatch,zhang2021flexmatch,anonymous2023softmatch}, increasing the risk of confirmation bias, we apply bias-corrected distillation-based consistency (hence high temperature) to uncertain samples, boosting model accuracy and reliability for LTSSL, as observed in Fig.~\ref{fig:semi_supervised_dists}.

\textbf{Imbalance-aware temperature selection.} Determining the optimal temperature for distillation can be critical, particularly when confronted with LT settings. While a temperature of $T$$=$$1$ is shown to be effective for balanced (unlabeled) datasets, see Fig.~\ref{fig:temperature-ablation}, this may not hold for imbalanced ones, as observed in fully-supervised settings~\cite{He2021DIVE_virtual_examples}. In response to this, we calculate the temperature to accommodate the class imbalance:
\begin{equation}
    T = \exp(\text{KL}(\mathcal{P}_{\text{bal}} \| \hat{Q})),
    \label{eq:temperature_inference}    
\end{equation}
by initiating distillation after a warm-up period. This temperature, once set, is kept constant throughout the remainder of the training. This strategy ensures that our distillation process is fine-tuned to the unlabeled data from the start and remains stable against minor changes in the pseudo-label distribution. We validate its effectiveness in Sec.~\ref{subsec:ablation_study}.

\section{Experiments}
\label{sec:experiments}

\subsection{Setup}

\textbf{Datasets.} We evaluate the performance of our approach on multiple benchmarks, assessing its robustness under varying degrees of class imbalance, labeled data availability, and class distribution mismatch.

\textbf{CIFAR10-LT} and \textbf{CIFAR100-LT} are based on CIFAR10 and CIFAR100 \cite{krizhevsky2009cifar}, each originally containing 60k 32$\times$32 color images across 10 and 100 classes, split into 50k for training and 10k for testing. Following the standard protocol \cite{kim2020darp}, we sample these datasets to create their long-tailed versions, using a head class size \( N_1 \) and an imbalance ratio \( \gamma_l \). $\gamma_u$ denotes the imbalanced ratio for unlabeled data. The number of images per class for labeled and unlabeled data is determined by \( N_k = N_1 \cdot \gamma_l^{-\kappa} \) and \( M_k = M_1 \cdot \gamma_{u}^{-\kappa} \), respectively, where $\kappa = {(k-1)/(K-1)}$. We use two labeled data settings with 1/3 and 1/9 of the total data \cite{oh2021daso}.

\textbf{STL10-LT} is derived by downsampling the labeled portion of the STL10 dataset \cite{coates2011stl10}, akin to the procedure used for the CIFAR long-tailed variants. STL10 consists of 5k training and 8k test images, each 96$\times$96 in resolution, spread across 10 classes. This is augmented by an extra 100k unlabeled images that include both in-distribution and related out-of-distribution (OOD) classes from the ImageNet \cite{deng2009imagenet} taxonomy.

\textbf{ImageNet127}~\cite{huh2016imagenet127}, a naturally imbalanced large-scale dataset with $\gamma_l \approx \gamma_u \approx 286$, groups the 1k classes of ImageNet \cite{deng2009imagenet} into 127 classes based on the WordNet hierarchy. We evaluate under 32$\times$32 and 64$\times$64 image resolution using 10\% of labeled data \cite{fan2021cossl}.

\textbf{Training and evaluation.} 
Our framework is based on FixMatch~\cite{sohn2020fixmatch} with a confidence threshold of 0.95~\cite{sohn2020fixmatch}. Our experiments use Wide-ResNet-28-2~\cite{zagoruyko2017wide} for CIFAR10-LT, CIFAR100-LT, and STL10-LT, while ResNet-50 is used for ImageNet-127, following ~\cite{kim2020darp,oh2021daso,fan2021cossl}. 
For CIFAR\{10,100\}-LT and STL10-LT, we train for 256 epochs of 1024 steps each, using SGD, Nesterov momentum of 0.9, and weight decay of 5e-4~\cite{oh2021daso}. The base learning rate (LR) is set to 0.03. ImageNet-127 experiments use Adam \cite{kingma2014adam} with a base LR of 0.002 for 500 epochs of 500 steps each, following~\cite{fan2021cossl}. Batch sizes are 64 for labeled and 128 for unlabeled data. We set \(\alpha_{\text{min}}\) to 0.1 and \(d\) to 2, following~\cite{lazarow2023udal}. A warm-up period of 50k steps is used with CIFAR\{10,100\}-LT, while STL10-LT and ImageNet127 skip it. Following common practices \cite{sohn2020fixmatch,kim2020darp,wei2021crest,lazarow2023udal}, we define equally weighted losses. An ablation study in Appendix A supports this choice. Appendix E includes pseudo-code for ADELLO and Appendix F details our hyperparameter settings.

To assess our method, an EMA network updates parameters at each step with a decay of 0.999 \cite{berthelot2019remixmatch,oh2021daso}. We report the average of the \textit{test balanced accuracy} over the final 20 epochs~\cite{fan2021cossl}. We provide the mean and standard deviation from three independent runs. Friedman ranking \cite{Friedman1937ranking, Friedman1940ranking} is used to fairly assess algorithms across different settings, subsequently determining the final ranking from the Friedman scores, following the methodology in \cite{wang2022usb}. All experiments were conducted on a single Nvidia V100-32GB GPU within a local cluster. A discussion on running times is deferred to Appendix B.

\subsection{Main Results}
\label{sec:main_results}
\begin{table}[t]
\centering
\caption{Test accuracy (\%) on CIFAR10-LT and CIFAR100-LT \textbf{under label shift}. $\dagger$: labeled prior as target. $\ddagger$: results from prior work \cite{oh2021daso}. Best scores \textbf{bold}, second-best \underline{underlined}.}
\label{tab:sota_cifar_more_labels}
\resizebox{0.75\linewidth}{!}{
\begin{tabular}{lcccccccc}
    \toprule
    & \multicolumn{3}{c}{CIFAR10-LT} & \multicolumn{3}{c}{CIFAR100-LT} & Friedman & Final \\
    \cmidrule(r){2-4}
    \cmidrule(r){5-7}
    \qquad  \ \ $\gamma_l$ $\rightarrow$ &  100 & 100 & 100  & 50 & 50 & 50 & Rank & Rank \\
    \cmidrule(r){8-9}
    \qquad  \ \ $\gamma_u$ $\rightarrow$ & 100 & 1 & 0.01 & 50 & 1 & 0.02 & & \\
    \qquad  \ \ $N_1$ $\rightarrow$ & 1500 & 1500 & 1500 & 150 & 150 & 150 & & \\
    \qquad  \ \ $M_1$ $\rightarrow$ & 3000 & 3000 & 30 & 300 & 300 & 6 & & \\
    \midrule
    Supervised & 63.8{\scriptsize $\pm$0.3} & 63.8{\scriptsize $\pm$0.3} & 63.8{\scriptsize $\pm$0.3}  & 36.3{\scriptsize $\pm$0.3} & 36.3{\scriptsize $\pm$0.3} & 36.3{\scriptsize $\pm$0.3} & - & - \\
    \midrule
    FixMatch \cite{sohn2020fixmatch} &  75.5{\scriptsize $\pm$1.1} & 86.1{\scriptsize $\pm$1.1} & 81.0{\scriptsize $\pm$4.2} & 44.4{\scriptsize $\pm$0.6} & 48.6{\scriptsize $\pm$1.0} & 45.4{\scriptsize $\pm$1.6} & 8.67 & 9 \\
    +DARP$^\dagger$\cite{kim2020darp} & 76.6{\scriptsize $\pm$1.0} & 68.8{\scriptsize $\pm$0.7} & 63.3{\scriptsize $\pm$1.3} & 44.7{\scriptsize $\pm$0.4} & 43.3{\scriptsize $\pm$0.6} & 40.4{\scriptsize $\pm$0.7} & 9.50 & 10 \\
    +CReST+ \cite{wei2021crest} & 78.1{\scriptsize $\pm$0.6} & \textbf{92.6}{\scriptsize $\pm$0.2} & 68.5{\scriptsize $\pm$0.5} & 44.9{\scriptsize $\pm$0.2}  & \underline{56.5}{\scriptsize $\pm$0.5} & 40.5{\scriptsize $\pm$0.4} & 6.17 & 7 \\  
    +ABC \cite{lee2021abc} & 82.3{\scriptsize $\pm$0.7} & 89.0{\scriptsize $\pm$0.2} & \textbf{87.0}{\scriptsize $\pm$0.4} & 47.2{\scriptsize $\pm$0.6} & 52.4{\scriptsize $\pm$1.5} & 48.7{\scriptsize $\pm$2.0} & 4.00 & 3 \\
    +DASO \cite{oh2021daso} & 79.1{\scriptsize $\pm$0.7}$^\ddagger$ & 88.8{\scriptsize $\pm$0.6}$^\ddagger$ & 80.3{\scriptsize $\pm$0.6}$^\ddagger$ & 44.7{\scriptsize $\pm$0.2} & 51.7{\scriptsize $\pm$2.0} & 48.5{\scriptsize $\pm$2.1} & 7.00 & 8 \\
    +DebiasPL \cite{wang2022debiasmatch} & 80.5{\scriptsize $\pm$0.1} & 88.6{\scriptsize $\pm$0.2} & 83.8{\scriptsize $\pm$0.2} & 46.8{\scriptsize $\pm$0.3} & 52.5{\scriptsize $\pm$0.8} & 50.8{\scriptsize $\pm$1.9} & 5.00 & 6 \\
    +CoSSL \cite{fan2021cossl} &  \textbf{84.6}{\scriptsize $\pm$0.1} & 88.8{\scriptsize $\pm$0.6} & 84.2{\scriptsize $\pm$0.2} & 47.6{\scriptsize $\pm$0.8} & 50.4{\scriptsize $\pm$1.2} & 46.8{\scriptsize $\pm$0.6} & 4.67 & 5 \\
    +UDAL$^\dagger$ \cite{lazarow2023udal} & 83.0{\scriptsize $\pm$0.3} & 89.1{\scriptsize $\pm$0.2} & 80.9{\scriptsize $\pm$0.7} & \underline{48.6}{\scriptsize $\pm$0.5} & 52.6{\scriptsize $\pm$1.0} & 48.7{\scriptsize $\pm$1.3}  & 4.00 & 3 \\
    \rowcolor{Gray}
    +ADELLO (ours) & \underline{83.8}{\scriptsize $\pm$0.3} & \underline{91.9}{\scriptsize $\pm$0.3} & \underline{86.1}{\scriptsize $\pm$0.4} & \textbf{49.2}{\scriptsize $\pm$0.6}     & \textbf{57.5}{\scriptsize $\pm$1.3}  & \textbf{53.0}{\scriptsize $\pm$0.9} & \textbf{1.50} & \textbf{1} \\
    \midrule
    SoftMatch \cite{anonymous2023softmatch} & 79.6{\scriptsize $\pm$0.2} & 89.6{\scriptsize $\pm$0.4} & 83.0{\scriptsize $\pm$0.8} & 46.4{\scriptsize $\pm$0.9}   &   \textbf{57.5}{\scriptsize $\pm$0.8}   &   \underline{51.2}{\scriptsize $\pm$1.2} & \underline{3.83} & \underline{2} \\
    \bottomrule
\end{tabular}
}
\end{table}

\begin{figure*}[tbp]
\begin{minipage}[t!]{0.65\linewidth}
\centering
\captionof{table}{Test accuracy (\%) on CIFAR\{10,100\}-LT and STL10-LT \textbf{under low-label regimes}. $\dagger$: labeled prior as target. $\ddagger$: results from prior work \cite{oh2021daso}. Best scores \textbf{bold}, second-best \underline{underlined}.}
\label{tab:sota_cifar_stl_fewer_labels}
\resizebox{\textwidth}{!}{%
\begin{tabular}{lcccccccc}
    \toprule
    & \multicolumn{2}{c}{CIFAR10-LT} & \multicolumn{2}{c}{CIFAR100-LT} & \multicolumn{2}{c}{STL10-LT} & Friedman & Final \\
    \cmidrule(r){2-3}
    \cmidrule(r){4-5}  
    \cmidrule(r){6-7}      
    \qquad \ \ $\gamma_l$ $\rightarrow$ & 100 & 150 & 10 & 20 & 10 & 20 & Rank & Rank \\
    \cmidrule(r){8-9}      
    \qquad \ \ $\gamma_u$ $\rightarrow$ & 100 & 150 & 10 & 20 & N/A & N/A \\
    \qquad \ \ $N_1$ $\rightarrow$ & 500 & 500 & 50 & 50 & 150 & 150 \\
    \qquad \ \ $M_1$ $\rightarrow$ & 4000 & 4000 & 400 & 400 & N/A & N/A \\
    \midrule
    Supervised & 46.6{\scriptsize $\pm$0.9} & 43.4{\scriptsize $\pm$1.9} & 27.7{\scriptsize $\pm$1.8} & 25.1{\scriptsize $\pm$1.1} & 46.4{\scriptsize $\pm$0.6} & 40.8{\scriptsize $\pm$0.6} & - & - \\
    \midrule
    FixMatch \cite{sohn2020fixmatch} & 69.8{\scriptsize $\pm$1.6} & 65.6{\scriptsize $\pm$1.5} & 47.0{\scriptsize $\pm$0.9} & 42.2{\scriptsize $\pm$0.6} & 64.1{\scriptsize $\pm$2.3} & 54.5{\scriptsize $\pm$4.3} & 9.67 & 10 \\
    +DARP$^\dagger$ \cite{kim2020darp} & 72.9{\scriptsize $\pm$1.3} & 67.2{\scriptsize $\pm$2.0} & 47.7{\scriptsize $\pm$0.7} & 42.8{\scriptsize $\pm$1.2} & 62.1{\scriptsize $\pm$1.4} & 54.7{\scriptsize $\pm$2.6} & 9.00 & 9 \\     
    +CReST+ \cite{wei2021crest} & 77.6{\scriptsize $\pm$0.2} & 72.1{\scriptsize $\pm$2.9} & 46.6{\scriptsize $\pm$0.6} & 43.2{\scriptsize $\pm$1.0} & 66.9{\scriptsize $\pm$1.0} & 62.6{\scriptsize $\pm$2.6}  & 7.33 & 8 \\
    +ABC \cite{lee2021abc} & 78.9{\scriptsize $\pm$0.9} & 72.0{\scriptsize $\pm$2.4} & 49.7{\scriptsize $\pm$1.3} & 44.1{\scriptsize $\pm$0.3} & 71.2{\scriptsize $\pm$1.0} & 65.7{\scriptsize $\pm$2.3} & 4.83 & 6 \\
    +DASO$^\ddagger$ \cite{oh2021daso} & 80.1{\scriptsize $\pm$1.2} & 70.6{\scriptsize $\pm$0.8} & 49.8{\scriptsize $\pm$0.2} & 43.6{\scriptsize $\pm$0.1} & 70.0{\scriptsize $\pm$1.2} & 65.7{\scriptsize $\pm$1.8} & 5.83 & 7 \\    
    +DebiasPL \cite{wang2022debiasmatch} & 76.4{\scriptsize $\pm$4.3} & 72.0{\scriptsize $\pm$1.8} & 50.3{\scriptsize $\pm$1.1} & \underline{45.4}{\scriptsize $\pm$0.5} & 70.1{\scriptsize $\pm$0.8} & 66.6{\scriptsize $\pm$2.1} & 4.50 & 4 \\
    +CoSSL \cite{fan2021cossl} & \underline{80.8}{\scriptsize $\pm$0.5} & \textbf{76.8}{\scriptsize $\pm$0.7} & 48.8{\scriptsize $\pm$1.0} & 44.4{\scriptsize $\pm$0.7} & 70.6{\scriptsize $\pm$0.5} & 66.0{\scriptsize $\pm$1.4} & 3.67 & 3 \\   
    +UDAL$^\dagger$ \cite{lazarow2023udal} & \underline{80.8}{\scriptsize $\pm$0.5} & \underline{76.4}{\scriptsize $\pm$2.6} & \underline{50.4}{\scriptsize $\pm$1.1} & \textbf{46.5}{\scriptsize $\pm$0.1} & 69.8{\scriptsize $\pm$1.1} & 65.0{\scriptsize $\pm$2.3} & \underline{3.50} & \underline{2} \\ 
    \rowcolor{Gray}
    +ADELLO (ours) & \textbf{81.3}{\scriptsize $\pm$0.4} & 76.0{\scriptsize $\pm$1.7} & \textbf{51.8}{\scriptsize $\pm$0.7} & \textbf{46.5}{\scriptsize $\pm$0.2} & \textbf{75.7}{\scriptsize $\pm$0.7} & \textbf{74.6}{\scriptsize $\pm$0.4} & \textbf{1.33} & \textbf{1} \\    
    \midrule
    SoftMatch \cite{anonymous2023softmatch} & 77.1{\scriptsize $\pm$0.8} & 71.0{\scriptsize $\pm$1.4} & 50.2{\scriptsize $\pm$0.7} & 43.8{\scriptsize $\pm$0.5} & \underline{72.6}{\scriptsize $\pm$0.3} & \underline{70.6}{\scriptsize $\pm$0.4} & 4.67 & 5 \\
    \bottomrule
\end{tabular}%
}
\end{minipage}
\hfill 
\begin{minipage}[t!]{0.34\linewidth}
    \begin{minipage}{\textwidth}
    \centering
    \captionof{table}{\textbf{Large-scale datasets}. Test balanced accuracy (\%) on ImageNet127 at 32 $\times$ 32 and 64 $\times$ 64 image resolution. $\dagger$: results from prior work \cite{fan2021cossl}. Best scores \textbf{bold}, second-best \underline{underlined}.}
    \label{table:large-scale-datasets}
    \resizebox{\textwidth}{!}{%
    \begin{tabular}{lcc}
    \toprule
    Method & 32 $\times$ 32 & 64 $\times$ 64 \\
    \midrule
    FixMatch \cite{sohn2020fixmatch}$^\dagger$ & 29.7 & 42.3 \\
    +DARP \cite{kim2020darp}$^\dagger$ & 30.5 & 42.5 \\
    +DARP +cRT \cite{kim2020darp}$^\dagger$ & 39.7 & 51.0 \\
    +CReST+ \cite{wei2021crest}$^\dagger$ & 32.5 & 44.7 \\
    +CReST+ +LA \cite{wei2021crest}$^\dagger$ & 40.9 & \underline{55.9} \\
    +CoSSL \cite{fan2021cossl}$^\dagger$ & 43.7 & 53.8 \\
    +UDAL ($\alpha_{\text{min}}$$=$$0.55$) \cite{lazarow2023udal} & 40.2 & 49.4 \\
    +UDAL ($\alpha_{\text{min}}$$=$$0.1$) \cite{lazarow2023udal} & \underline{44.1} & 52.3 \\
    \rowcolor{Gray}
    +ADELLO (ours) & \textbf{47.5} & \textbf{58.0}  \\
    \bottomrule
    \end{tabular}%
    }
    \end{minipage}
\end{minipage}
\end{figure*}



In this section, we present extensive experiments to evaluate the performance of our approach against several SOTA approaches. These methods include a supervised baseline (using only labeled data), FixMatch \cite{sohn2020fixmatch} (SSL baseline), SoftMatch \cite{anonymous2023softmatch} (stronger SSL baseline), as well as representative LTSSL algorithms, including DARP \cite{kim2020darp}, CReST+ \cite{wei2021crest}, ABC \cite{lee2021abc}, DASO \cite{oh2021daso}, DebiasPL \cite{wang2022debiasmatch}, CoSSL \cite{fan2021cossl}, and UDAL \cite{lazarow2023udal}. To demonstrate the effectiveness of our method across diverse scenarios, we assess its performance under varying levels of label shift in Table~\ref{tab:sota_cifar_more_labels}, different degrees of class imbalance under low-label regimes in Table~\ref{tab:sota_cifar_stl_fewer_labels}, and an exceptionally challenging scenario on ImageNet127 in Table~\ref{table:large-scale-datasets}. Furthermore, we investigate model calibration in Section~\ref{sec:calibration}.

\textbf{Results under (unknown) label shift.}
To address scenarios where the unlabeled class distribution differs from or is unknown relative to the labeled prior, we vary the unlabeled class distribution for CIFAR\{10,100\}-LT to obtain three evaluation settings: forward long-tailed ($\gamma_u=\gamma_l$), balanced ($\gamma_u=\frac{1}{K}$), and reversed long-tailed ($\gamma_u=\frac{1}{\gamma_l}$).

Table~\ref{tab:sota_cifar_more_labels} demonstrates the robustness of ADELLO in handling unknown distribution mismatches, particularly evident on CIFAR100-LT (see also Fig.~\ref{fig:label_shift_top}), which contains a larger number of classes. The rightmost columns of the table show the Friedman scoring and the final rank over test accuracies for each method. ADELLO secures the top position, showcasing its superior performance. Consistently ranking first or second in all settings, it demonstrates remarkable adaptability to degrees of label shift. The effectiveness of ADELLO becomes evident in both forward and reversed LT scenarios on CIFAR100, outperforming SoftMatch significantly. Distinctly outperforming previous SOTA approaches like ABC, DASO, and CoSSL, ADELLO delivers robust LTSSL performance without depending on auxiliary classifiers or data re-sampling.

\begin{wrapfigure}{r}{0.27\linewidth} 
    \vspace{-20pt}       
    \centering
    \includegraphics[trim={0.3cm 0.3cm 0.3cm 0.2cm},clip,width=\linewidth]{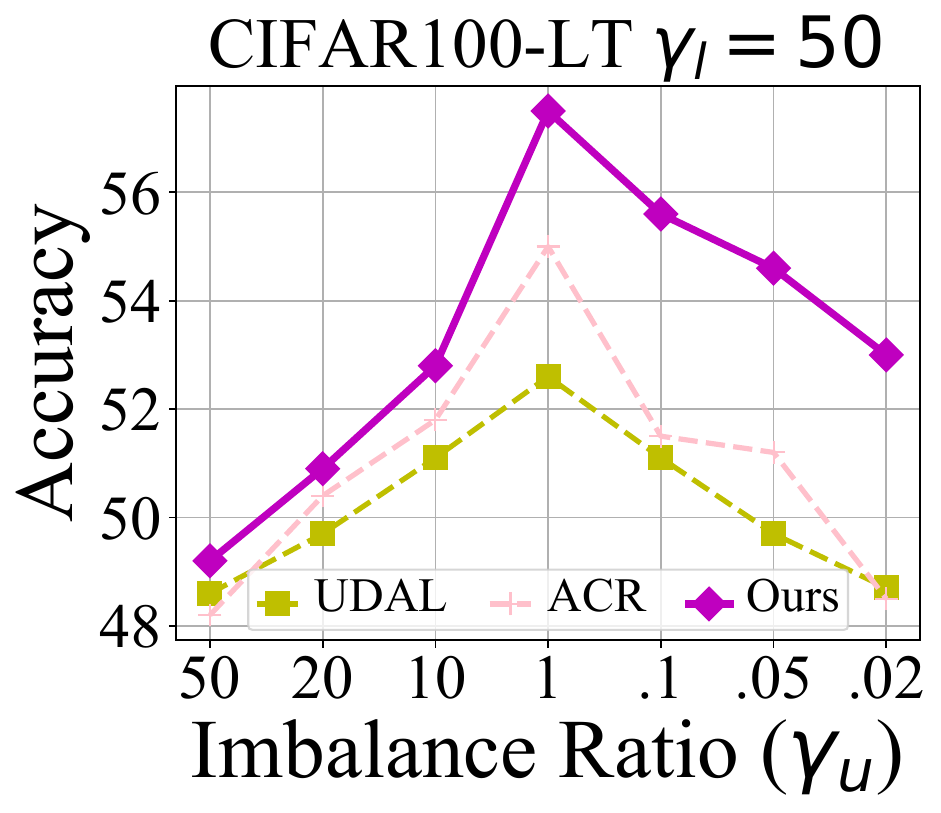}
    \vspace{-16pt}
    \caption{Varying label shift.}
    \label{fig:dist_shift}
    \vspace{-20pt}
 \end{wrapfigure}

We also compare the classification performance of ADELLO with ACR \cite{wei2023acr}, a recent LTSSL approach. Fig.~\ref{fig:dist_shift} shows that ADELLO outperforms ACR under label shift, with the performance gap widening as the distribution mismatch increases.

\textbf{Results with limited labeled data.} Table~\ref{tab:sota_cifar_stl_fewer_labels} highlights the effectiveness of ADELLO on CIFAR\{10,100\}-LT and STL10-LT, particularly in scenarios with limited labeled data and significant class imbalance. In STL10-LT, where only 150 labels are available for the head class amid a range of OOD data, ADELLO shows marked improvements. It notably surpasses CoSSL with a +8.0 gain in accuracy and ABC with a +4.5 increase at imbalance ratios of 20 and 10, respectively, while outperforming SoftMatch, a strong SSL baseline. Oddly, SoftMatch mistakenly classifies OOD data as known classes using hard PLs with targeted weights. Conversely, our method uses CCR to predict soft PLs on potential OOD samples, enhancing robustness. 

Under consistent CIFAR\{10,100\}-LT settings, the performance of ADELLO matches or exceeds that observed in established methods like CoSSL and UDAL, reinforcing the effectiveness of correct prior estimation even with a low amount of labels and without reliance on strong assumptions. Notably, ADELLO outperforms SoftMatch by a large margin as imbalance ratios increase without using any adaptive thresholding technique.

\textbf{Results on ImageNet127.} The performance of various methods on the ImageNet127 dataset, a challenging variant of standard ImageNet featuring 127 classes and an imbalance ratio of 286, is summarized in Table~\ref{table:large-scale-datasets}. Due to its extensive sample size (1.28M), ImageNet127 serves as a unique testbed for assessing large-scale imbalanced datasets. ADELLO notably excels in balanced accuracy, surpassing the previous state-of-the-art, CoSSL, with gains of +3.8 at 32$\times$32 resolution and +4.2 at 64$\times$64 resolution. Compared to UDAL, accuracy increases of +3.4 and +5.7 are observed, respectively, at these resolutions. The advantage of ADELLO over UDAL and CReST+, both using consistent priors for distribution alignment, highlights the benefits of marrying FlexDA with complementary consistency for superior performance in large-scale settings.

\subsection{Study of Model Calibration}
\label{sec:calibration}
\begin{figure*}[tbp]
\centering
\begin{subfigure}{.75\textwidth}
  \centering
  \includegraphics[width=\linewidth]{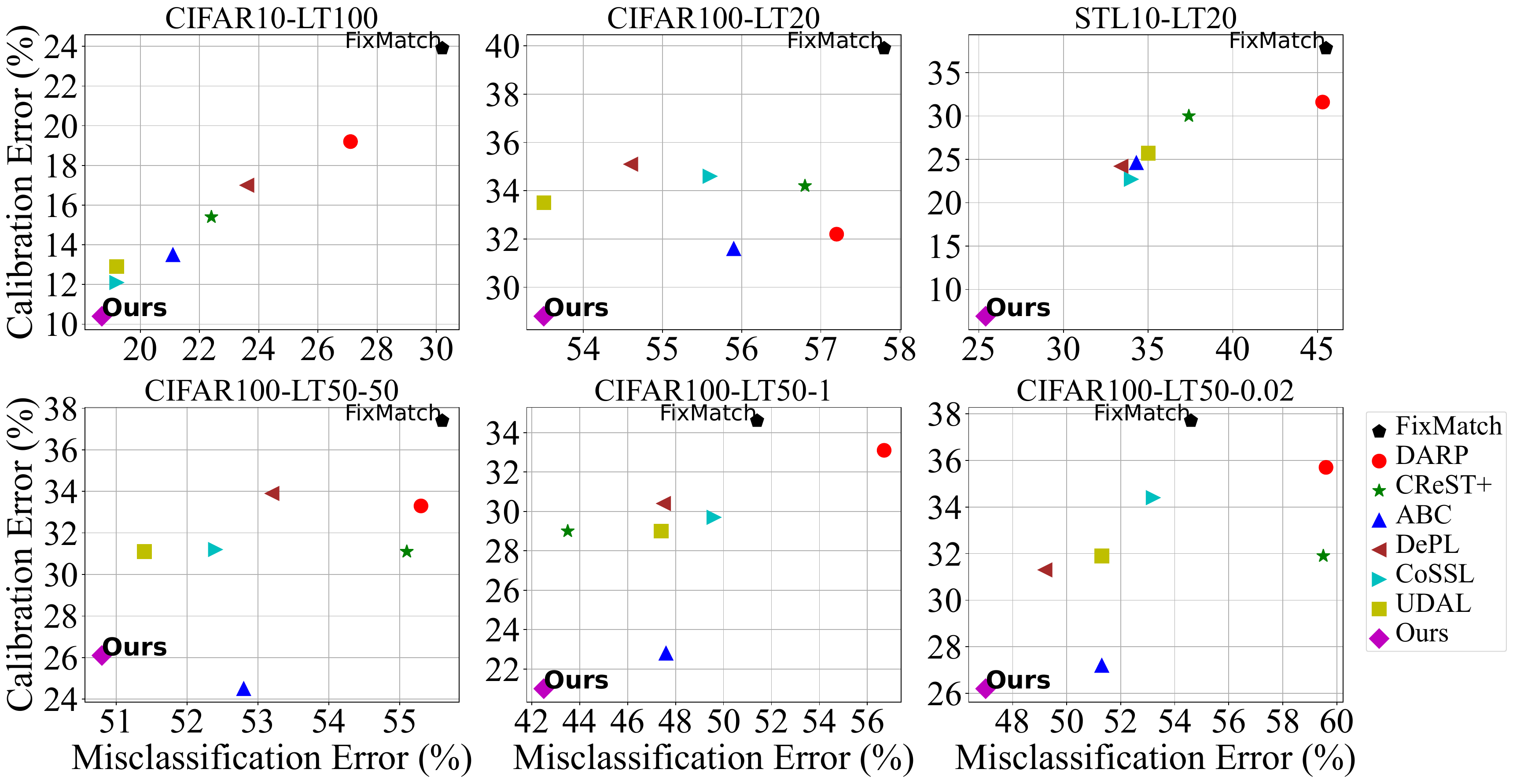}
\end{subfigure}%
\caption{\textbf{Trade-off between Generalization and Calibration Performance}. We report misclassification error (\%) vs. expected calibration error (\%), computed on the test split. The first row contrasts different datasets, and the second row examines various degrees of label shift.}
\label{fig:generalization_vs_calibration}
\end{figure*}

Model calibration~\cite{guo2017calibration,loh2022calibrationssl}, vital for accurately reflecting predictive uncertainty in SSL, is a focal point of our study~\cite{loh2022calibrationssl}. We examine the impact of calibration on LTSSL by comparing expected calibration error (ECE) with misclassification error. Our findings align with \cite{loh2022calibrationssl}, showing that better-calibrated models improve SSL performance, even with class imbalances, as Fig.~\ref{fig:generalization_vs_calibration} illustrates. ADELLO demonstrates a superior trade-off in reducing misclassification and calibration errors compared to other LTSSL methods, across CIFAR\{10,100\}-LT and STL10-LT datasets. Fig.~\ref{fig:generalization_vs_calibration} (first row) reveals that while most LTSSL methods surpass FixMatch, ADELLO further improves generalization and, particularly, calibration in scenarios where distributions align. The superiority of ADELLO is significantly highlighted in the STL10-LT benchmark, characterized by an unknown label shift and a substantial presence of unlabeled OOD samples. Section~\ref{subsec:ablation_study} attributes these improvements to our flexible distribution alignment and the significant role of complementary consistency regularization in such contexts.

In scenarios with controlled label shift, as illustrated in Fig.~\ref{fig:generalization_vs_calibration} (second row), certain LTSSL methods, such as DARP, CoSSL, and CReST+ to a degree, face difficulties with large label shifts where the bias in labeled data fails to accurately represent the characteristics of the unlabeled distribution, a problem that ADELLO overcomes. Although the auxiliary balanced classifier in ABC demonstrates acceptable calibration, ADELLO showcases greater flexibility and robustness in generalization performance compared to ABC. Appendix C presents similar trends for the maximum calibration error (MCE).

\subsection{Ablation Study}
\label{subsec:ablation_study}
\begin{figure*}[tbp]
\begin{minipage}[t!]{0.49\linewidth}
    \centering
    \captionof{table}{Influence of ADELLO components on model generalization (Test accuracy).}
    \label{tab:ablation_cifar}
    \resizebox{\textwidth}{!}{
    \begin{tabular}{lcccc}
        \toprule
        Components & \multicolumn{3}{c}{CIFAR100-LT50 $\uparrow$} & \multicolumn{1}{c}{STL10-LT20 $\uparrow$}  \\       
        \cmidrule(r){1-1}   
        \cmidrule(r){2-4}
        \cmidrule(r){5-5}           
         \qquad $\gamma_u$ $\rightarrow$ & 50 & 1 & 0.02 & N/A \\        
        \midrule
        FixMatch & 44.4{\scriptsize $\pm$0.6} & 48.6{\scriptsize $\pm$1.0} & 45.4{\scriptsize $\pm$1.6} & 54.5{\scriptsize $\pm$4.3} \\
        +FlexDA & 48.6{\scriptsize $\pm$0.7} & 53.6{\scriptsize $\pm$1.0} & 51.2{\scriptsize $\pm$0.9} & 67.1{\scriptsize \(\pm\)1.6} \\
        +CCR & 44.7{\scriptsize $\pm$0.7} & 51.6{\scriptsize $\pm$1.6} & 47.2{\scriptsize $\pm$2.1} & 61.1{\scriptsize \(\pm\)2.9} \\  
        \rowcolor{Gray}
        +\textbf{FlexDA+CCR} & \textbf{49.2}{\scriptsize $\pm$0.6} & 57.5{\scriptsize $\pm$1.3}  & \textbf{53.0}{\scriptsize $\pm$0.9} & \textbf{74.6}{\scriptsize \(\pm\)0.4} \\
       \midrule        
        +FlexDA+KD & 49.1{\scriptsize $\pm$0.6} & \textbf{58.2}{\scriptsize $\pm$1.1} & 52.8{\scriptsize $\pm$1.1} & 74.4{\scriptsize \(\pm\)0.5} \\        
        \bottomrule        
    \end{tabular}
    }
\end{minipage}
\hfill 
\begin{minipage}[t!]{0.48\linewidth}
    \centering
    \captionof{table}{Influence of ADELLO components on model calibration (ECE/MCE).}
    \label{tab:ablation_calibration}
    \resizebox{\textwidth}{!}{
    \begin{tabular}{lcc}
    \toprule
        Components & \multicolumn{1}{c}{CIFAR100-LT50 $\downarrow$} & \multicolumn{1}{c}{STL10-LT20 $\downarrow$}  \\ 
        \cmidrule(r){1-1}   
        \cmidrule(r){2-2}
        \cmidrule(r){3-3}                  
         \qquad $\gamma_u$ $\rightarrow$ & 50 & N/A \\        
        \midrule        
        FixMatch & 37.4{\scriptsize $\pm$0.4} / 57.3{\scriptsize $\pm$1.1} & 37.8{\scriptsize $\pm$4.5} / 55.1{\scriptsize $\pm$4.9} \\
        +FlexDA & 31.4{\scriptsize $\pm$0.4} / 52.0{\scriptsize $\pm$2.4} & 23.6{\scriptsize $\pm$1.4} / 49.5{\scriptsize $\pm$4.5} \\
        +CCR & 36.3{\scriptsize $\pm$0.7} / 56.8{\scriptsize $\pm$1.8} & 22.2{\scriptsize $\pm$2.3} / 38.5{\scriptsize $\pm$4.2} \\  
        \rowcolor{Gray}
        +\textbf{FlexDA+CCR} & \textbf{26.1}{\scriptsize $\pm$0.9} / \textbf{46.2}{\scriptsize $\pm$0.6} & \textbf{6.9}{\scriptsize $\pm$0.3} / \textbf{25.9}{\scriptsize $\pm$1.0} \\
        \midrule  
        +FlexDA+KD & 33.4{\scriptsize $\pm$0.6} / 57.5{\scriptsize $\pm$2.0} & 10.0{\scriptsize $\pm$0.5} / 31.3{\scriptsize $\pm$7.5} \\   
        \bottomrule  
    \end{tabular}
    }
\end{minipage}
    \centering
    \begin{minipage}[t]{0.22\linewidth} 
        \centering
        \captionof{table}{Ablation of scheduler speed (\(d\)).}
        \label{table:schedule_ablation}
        \begin{tabular}{cc}
        \toprule
        d  & $\gamma_u = 50$ \\
        \midrule
        0 & 46.6{\scriptsize $\pm$0.6} \\
        1 & 49.1{\scriptsize $\pm$0.4} \\
        2 & 49.2{\scriptsize $\pm$0.5} \\
        3 & 49.1{\scriptsize $\pm$0.8} \\
        \bottomrule
        \end{tabular}
    \end{minipage}%
    \hfill 
    \begin{minipage}[t]{0.22\linewidth} 
        \centering
        \captionof{table}{Ablation of minimum bias (\(\alpha_{\text{min}}\)).}
        \label{table:min_bias_ablation}
        \begin{tabular}{cc}
        \toprule
        $\alpha_{\text{min}}$ & $\gamma_u = 50$ \\
        \midrule
        0.0 & 49.1{\scriptsize $\pm$0.7} \\
        0.1 & 49.2{\scriptsize $\pm$0.5} \\
        0.2 & 49.1{\scriptsize $\pm$1.2} \\
        0.3 & 48.7{\scriptsize $\pm$0.7} \\
        \bottomrule
        \end{tabular}
    \end{minipage}
    \hfill 
    \begin{minipage}[t]{0.48\linewidth} 
        \centering
        \captionof{table}{Ablation of warm-up period.}
        \label{table:warmup_period_ablation}
            \begin{tabular}{lccc}
            \toprule
            \#steps \qquad $\gamma_u$ $\rightarrow$ & 50 & 1 & 0.02 \\
            \midrule
            no warm-up & 45.6{\scriptsize $\pm$0.6} & 58.4{\scriptsize $\pm$1.3} & 51.3{\scriptsize $\pm$1.9} \\
            25k steps & 49.2{\scriptsize $\pm$0.7} & 57.7{\scriptsize $\pm$1.3} & 52.6{\scriptsize $\pm$1.0} \\
            50k steps & 49.2{\scriptsize $\pm$0.6} & 57.5{\scriptsize $\pm$1.3} & 53.0{\scriptsize $\pm$0.9} \\
            100k steps & 49.1{\scriptsize $\pm$0.4} & 57.8{\scriptsize $\pm$1.2} & 52.8{\scriptsize $\pm$1.0} \\
            no distillation & 48.6{\scriptsize $\pm$0.7} & 53.6{\scriptsize $\pm$1.0} & 51.2{\scriptsize $\pm$0.9} \\
            \bottomrule
    \end{tabular}
    \end{minipage}
\end{figure*}

\textbf{Importance of the proposed losses.}\label{comp} Our ablation studies on CIFAR100-LT50, shown in Table \ref{tab:ablation_cifar}, evaluate ADELLO objectives across imbalance ratios ($\gamma_u$) of 50, 1, and 0.02. The FlexDA component in ADELLO significantly outperforms the baseline, FixMatch, with gains of +4.2, +5.0, and +5.8 points for these \(\gamma_u\) values, underscoring its effectiveness against class imbalance and distribution mismatch. Further, the studies indicate that complementary consistency enhances performance, highlighting its value in SSL. However, the synergy of FlexDA and CCR within ADELLO results in the most substantial improvements, with increases of +4.8, +8.9, and +7.6 points across the 50, 1, and 0.02 imbalance ratios, respectively. 
While FlexDA sees advantages from indiscriminate KD of all samples (FlexDA+KD), using masked distillation (FlexDA+CCR) more often results in enhanced generalization in imbalanced scenarios (see $\gamma_u \in \{50,0.02\}$).

\textbf{Are all components in ADELLO necessary for proper calibration?} Table~\ref{tab:ablation_calibration} shows that both FlexDA and CCR boost calibration independently. We observe that their synergy markedly surpasses the baseline, by correcting the label bias on the whole data distribution, akin to fully-supervised approaches \cite{xu2021bayias_calibrated,aimar2023balpoe_calibrated}. Significantly, the key to enhanced calibration in LTSSL contexts lies not just in the naive distillation of all samples (FlexDA+KD), but in the strategic combination of soft pseudo-labels for underconfident samples and hard pseudo-labels for those with high confidence, as depicted in Fig.~\ref{fig:adello-diagram}.

\textbf{Do we need a progressive scheduler?} The setup for FlexDA, as outlined in Section~\ref{sec:main_results}, adheres to configurations proposed by Lazarow et al.~\cite{lazarow2023udal}. Within the ADELLO framework, this analysis investigates the effect of the speed of the scheduler (\(d\)) and the minimum bias hyperparameters (\(\alpha_{\text{min}}\)) on model performance. Table~\ref{table:schedule_ablation} suggests that a moderate, yet progressive, scheduler, i.e. $d \in (1,3)$, leads to optimal accuracy, while an aggressive debiasing rate (\(d = 0\)) proves detrimental. Performance peaks when the minimum bias (\(\alpha_{\text{min}}\)) is near zero as observed in Table~\ref{table:min_bias_ablation}, suggesting that this configuration minimizes the balanced error, aligning with findings of fully-supervised approaches~\cite{menon2021longtail}. Generally, there is minimal sensitivity to the precise settings of these hyperparameters. We hypothesize that the union of FlexDA with complementary consistency is key to the effectiveness of ADELLO compared to other DA approaches, namely CReST+ and DebiasPL, which do not engage in full classifier debiasing to retain high data utilization.

\begin{wrapfigure}{r}{0.27\linewidth} 
    \vspace{-20pt}
    \centering
    \includegraphics[width=\linewidth]{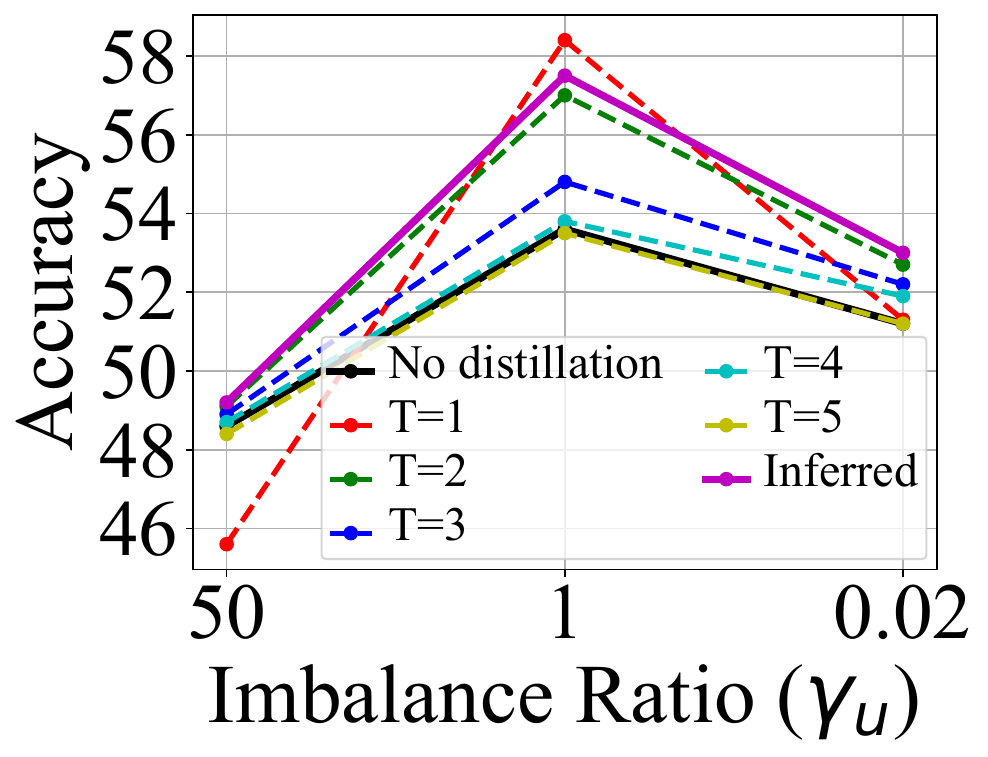}
    \vspace{-20pt}
    \caption{Inferred vs. tuned temperature.} \label{fig:temperature-ablation}
    \vspace{-20pt}
 \end{wrapfigure} 
 
\textbf{How robust is the inferred temperature $T$?} We calibrate $T$ based on the class imbalance of unlabeled data, using (\ref{eq:temperature_inference}). Fig.~\ref{fig:temperature-ablation} shows 
this strategy is nearly as effective as custom-tuning the temperature for each dataset. A $T$ near one is preferred for balanced data, facilitating distillation from an increasingly balanced classifier and leading to marked performance gains. Under more imbalanced cases, our method takes a more cautious approach by opting for a higher temperature.

\textbf{Effect of warm-up period.} Table~\ref{table:warmup_period_ablation} shows that beginning complementary consistency after a warm-up stage using our temperature-selection procedure boosts performance compared to using an uninformative setting or not distilling at all, while showing robustness to the exact starting point for CCR.

\section{Conclusion}
\label{sec:conclusions}

We proposed a two-faceted framework for greatly improving the performance of LTSSL under label shift. First, our flexible distribution alignment (FlexDA) reduces the bias caused by differing labeled and unlabeled class-distribution marginals, and subsequently, the head-class bias intrinsic to imbalanced data. These reductions are achieved by aligning the model prior first to a dynamic estimate of the unlabeled marginal and gradually towards a more balanced distribution. Second, our complementary consistency regularization leverages the soft output signals of below-threshold pseudo-labels toward improving data utilization of minority classes. We demonstrate that this framework is state-of-the-art when unlabeled and labeled marginal distributions are mismatched, competitive when they are matched, and achieves better calibration than its competitors.

\textbf{Limitations.}
All LTSSL benchmarks we are aware of focus on the \textit{closed-world assumption} \cite{closed_world_assumption}, where every class is labeled and known during inference. For the STL10-LT, which includes near-out-of-distribution unlabeled samples, ADELLO yields promising results, suggesting potential in handling "unknown" classes not present in the labeled set. Furthermore, while the proposed framework is developed for the classification task, it may also be beneficial to address the class imbalance in more complex visual tasks, such as object detection, instance segmentation, or tracking.

\section*{Acknowledgements}
This work was supported by the Wallenberg Artificial Intelligence, Autonomous Systems and Software Program (WASP), funded by the Knut and Alice Wallenberg Foundation. The computational resources were provided by the National Academic Infrastructure for Supercomputing in Sweden (NAISS), partially funded by the Swedish Research Council through grant agreement no. 2022-06725, and by the Berzelius resource, provided by the Knut and Alice Wallenberg Foundation at the National Supercomputer Centre.

%
%
\bibliographystyle{splncs04}
\bibliography{main}

\clearpage
\newpage
\appendix

\appendix
\setcounter{table}{9}
\setcounter{figure}{6}


\begin{center}\textbf{\Large Supplementary Material}
\end{center}


The appendix includes the following sections:
\begin{enumerate}
    \item \textbf{Simplified objective} (Appendix~\ref{sec:simplified_objective}): discusses the proposed objective parameterization.
    \item \textbf{Computational efficiency} (Appendix~\ref{sec:algorithmic_complexity}): discusses complexity and training speed of ADELLO.
    \item \textbf{Confidence calibration} (Appendix~\ref{sec:extended_discussion_calibration}): provides additional definitions and further discussion about calibration performance.
    
    \item \textbf{Beyond natural images} (Appendix~\ref{sec:additional_domains}): presents experiments conducted on additional image domains, including medical and remote sensing datasets.
    
    \item \textbf{Additional algorithmic details} (Appendix~\ref{sec:adello_algorithm}): includes pseudo-code of the proposed algorithm.
    \item \textbf{Additional training details} (Appendix~\ref{sec:hyperparameters_settings}): includes hyperparameter configurations for each dataset.
\end{enumerate}

\section{Simplified objective}
\label{sec:simplified_objective}
In the main paper, we utilize equally weighted losses for ADELLO. Alternatively, a more complex formulation can be expressed as:
\begin{equation}
\mathcal{L} = \mathcal{L}_s^{\text{FlexDA}} + \lambda_{u} \mathcal{L}_{u}^{\text{FlexDA}} + \lambda_{uC} \mathcal{L}_{uC}^{\text{FlexDA}}\,,
\end{equation}
where $\lambda_u$ and $\lambda_{uC}$ are loss weights assigned to standard consistency and complementary consistency losses within the FlexDA framework, respectively. 
For simplicity and following the accepted $\lambda_{u} = 1$ norm \cite{kim2020darp,lazarow2023udal,sohn2020fixmatch,wei2021crest}, we also set $\lambda_{uC} = 1$. Table~\ref{table:lambda_u_ablation_cifar} supports this choice across several datasets, presenting steady performance around the default setting, with a decline noted for extreme values.
\begin{table}[h]
\centering
\caption{Ablation of complementary consistency loss weight \(\lambda_{uC}\). We report test accuracy using CIFAR100-LT50 and STL10-LT20 datasets.}
\label{table:lambda_u_ablation_cifar}
\resizebox{0.95\textwidth}{!}{
\begin{tabular}{ccccccccccc}
\toprule
\(\lambda_{uC}\)\ & 0 & 0.001 & 0.01 & 0.1 & 0.5 & \textbf{1} & 2 & 10 \\
\midrule
CIFAR100-LT50 & 48.6{\scriptsize \(\pm\)0.7} & 48.4{\scriptsize \(\pm\)0.7} & 48.6{\scriptsize \(\pm\)1.0} & 48.8{\scriptsize \(\pm\)0.6} & 49.0{\scriptsize \(\pm\)0.5} & \textbf{49.2}{\scriptsize \(\pm\)0.5} & 48.5{\scriptsize \(\pm\)0.5} & 44.5{\scriptsize \(\pm\)0.5}  \\
STL10-LT20 & 67.1{\scriptsize \(\pm\)1.6} & 67.3{\scriptsize \(\pm\)1.4} & 67.4{\scriptsize \(\pm\)1.1} & 69.3{\scriptsize \(\pm\)1.0} & 74.0{\scriptsize \(\pm\)0.6} & \textbf{74.6}{\scriptsize \(\pm\)0.4} & 72.4{\scriptsize \(\pm\)1.3} & 71.4{\scriptsize \(\pm\)0.3}  \\
\bottomrule
\end{tabular}
}
\end{table}

\section{Computational efficiency}
\label{sec:algorithmic_complexity}
ADELLO improves FixMatch by aligning pseudo-labels with the (unknown) class distribution of unlabeled data. This is achieved by tracking the exponential moving average of pseudo-labels, which is then used to adjust cross-entropy losses to correct for long-tailed biases. Additionally, it employs a masked distillation loss. Importantly, ADELLO accomplishes these enhancements without increased complexity. It does so by avoiding additional computational steps such as extra forward passes, the use of auxiliary classifiers, or the need for data re-sampling, thus maintaining a straightforward implementation. Training times show its efficiency: \textbf{ADELLO} at \textbf{5h18m} closely aligns with \textbf{FixMatch} at \textbf{5h15m} and ABC at 5h21m, and surpasses CReST+ at 6h22m, CoSSL at 7h29m, DARP at 7h43m, and \textbf{DASO} at \textbf{19h32m} for CIFAR100-LT50 on a single Nvidia V100-32GB GPU.

\section{Confidence calibration} 
\label{sec:extended_discussion_calibration}

\textbf{Calibration definitions.} At its core, model calibration evaluates how closely a model's predicted confidence aligns with the actual likelihood of correctness~\cite{Brocker2008ReliabilitySA}. For example, if a model predicts a certain class with 95\% confidence, in an ideal scenario, that prediction should be accurate 95\% of the time. A practical calibration requirement is \textit{argmax calibration}~\cite{guo2017calibration}. For a model $P$, outputting normalized probabilities, this criterion requires that for the class with the highest predicted confidence, denoted as $\hat{Y} = \argmax P(X)$ with confidence $\hat{P}(X) = \max P(X)$, said confidence should match the actual probability of that class being correct, across all levels of confidence:
\begin{equation}
    \mathbb{P}(\hat{Y} = Y | \hat{P}(X) = p ) \overset{!}{=} p, \ \ \ \ \ \  \forall p \in [0, 1].
    \label{eqn:argmax_calibration}
\end{equation}
In practice, we empirically evaluate the congruence between predicted confidence and actual accuracy over a test dataset $\mathcal{D}_{\text{test}} = \{x_i, y_i\}^{N_\text{test}}_{i=1}$. This involves grouping model predictions into $M$ bins based on confidence levels and analyzing the accuracy and confidence within each bin. For a given bin $B_m$, its accuracy, $\mathit{acc}(B_m)$, is the proportion of correct predictions, and its confidence, $\mathit{conf}(B_m)$, is the average predicted confidence. The Expected Calibration Error (ECE) quantifies the overall discrepancy between accuracy and confidence across all bins:
\begin{equation}
    ECE = \sum_{m=1}^{M} \frac{|B_m|}{n} |\mathit{acc}(B_m) - \mathit{conf}(B_m)|.
    \label{eqn:ece_metric}
\end{equation}
Similarly, the Maximum Calibration Error (MCE) identifies the largest such discrepancy, indicating the worst-case deviation between confidence and accuracy:
\begin{equation}
    MCE = \max_{m \in \{1,..,M\}} |\mathit{acc}(B_m) - \mathit{conf}(B_m)|.
    \label{eqn:mce_metric}
\end{equation}

\textbf{More results on model calibration.} 
In 
Section~5.3,
we show how our approach not only improves generalization capabilities but also significantly enhances model calibration in various LTSSL contexts. Additionally, in Tables~\ref{tab:ece_results_reordered} and \ref{tab:mce_results_reordered}, we present the calibration performance of various models, focusing specifically on the expected calibration error and the maximum calibration error, respectively. Our approach is consistently the top performer for reducing ECE, as shown in Table~\ref{tab:ece_results_reordered}. Analogously, ADELLO achieves the leading position for MCE reduction, as presented in Table~\ref{tab:mce_results_reordered}. This aspect is especially crucial in mission-critical applications, where reducing the maximum errors in model predictions is imperative.


\begin{table}[htbp]
\centering
\caption{\textbf{Expected Calibration Error (ECE) across different datasets.} Best scores \textbf{bold}, second-best \underline{underlined}.}
\label{tab:ece_results_reordered}
\resizebox{0.9\textwidth}{!}{
\begin{tabular}{lcccccccc}
\toprule
& \multicolumn{1}{c}{CIFAR10-LT} & \multicolumn{1}{c}{STL10-LT} & \multicolumn{4}{c}{CIFAR100-LT} & Friedman & Final \\
\cmidrule(r){2-2}
\cmidrule(r){3-3}  
\cmidrule(r){4-7}      
\qquad \ \ $\gamma_l$ $\rightarrow$ & 100 & 20 & 20 & 50 & 50 & 50 & Rank & Rank \\
\cmidrule(r){8-9}      
\qquad \ \ $\gamma_u$ $\rightarrow$ & 100 & N/A & 20 & 50 & 1 & 0.02 \\
\qquad \ \ $N_1$ $\rightarrow$ & 500 & 150 & 50 & 150 & 150 & 150 \\
\qquad \ \ $M_1$ $\rightarrow$ & 4000 & N/A & 400 & 300 & 300 & 6 \\
\midrule
FixMatch \cite{sohn2020fixmatch} & 23.9{\scriptsize $\pm$1.9} & 37.8{\scriptsize $\pm$4.5} & 39.9{\scriptsize $\pm$0.4} &  37.4{\scriptsize $\pm$0.4} & 34.6{\scriptsize $\pm$0.5} & 37.7{\scriptsize $\pm$0.8} & 9.0 & 9 \\
+DARP \cite{kim2020darp} & 19.2{\scriptsize $\pm$1.2} & 31.6{\scriptsize $\pm$2.9} & 32.2{\scriptsize $\pm$0.5} & 33.3{\scriptsize $\pm$0.1} & 33.1{\scriptsize $\pm$0.7} & 35.7{\scriptsize $\pm$0.7} & 6.8 & 8 \\
+CReST+ \cite{wei2021crest} & 15.4{\scriptsize $\pm$0.3} & 30.0{\scriptsize $\pm$2.7} & 34.2{\scriptsize $\pm$0.7} & 31.1{\scriptsize $\pm$0.1} & 29.0{\scriptsize $\pm$0.7} & 31.9{\scriptsize $\pm$0.8}  & 5.1 & 5 \\
+ABC \cite{lee2021abc} & 13.5{\scriptsize $\pm$1.0} & 24.6{\scriptsize $\pm$2.3} & \underline{31.6}{\scriptsize $\pm$0.2} & \textbf{24.5}{\scriptsize $\pm$0.5} & \underline{22.8}{\scriptsize $\pm$0.7} & \underline{27.2}{\scriptsize $\pm$1.5} & \underline{2.7} & \underline{2} \\
+DebiasPL \cite{wang2022debiasmatch} & 17.0{\scriptsize $\pm$4.2} & 24.2{\scriptsize $\pm$1.1} & 35.1{\scriptsize $\pm$1.1} & 33.9{\scriptsize $\pm$0.3} & 30.4{\scriptsize $\pm$0.7} & 31.3{\scriptsize $\pm$1.3} & 5.8 & 7 \\
+CoSSL \cite{fan2021cossl} & \underline{12.1}{\scriptsize $\pm$0.5} & 22.7{\scriptsize $\pm$1.3} & 34.6{\scriptsize $\pm$0.5} & 31.2{\scriptsize $\pm$0.3} & 29.7{\scriptsize $\pm$0.9} & 34.4{\scriptsize $\pm$0.7} & 4.8 & 4 \\
+UDAL \cite{lazarow2023udal} & 12.9{\scriptsize $\pm$0.4} & 25.7{\scriptsize $\pm$2.3} & 33.5{\scriptsize $\pm$0.3} & 31.1{\scriptsize $\pm$0.1} & 29.0{\scriptsize $\pm$0.7} & 31.9{\scriptsize $\pm$0.8} & 4.4 & 3 \\

\rowcolor{Gray}
+ADELLO (ours) & \textbf{10.4}{\scriptsize $\pm$0.3} & \textbf{6.9}{\scriptsize $\pm$0.3} & \textbf{28.8}{\scriptsize $\pm$0.3} & \underline{26.1}{\scriptsize $\pm$0.9} & \textbf{21.0}{\scriptsize $\pm$0.9} & \textbf{26.2}{\scriptsize $\pm$0.5} & \textbf{1.2} & \textbf{1} \\
\midrule
SoftMatch \cite{anonymous2023softmatch} & 15.7{\scriptsize $\pm$0.8} & \underline{20.0}{\scriptsize $\pm$0.5} & 36.7{\scriptsize $\pm$0.3} & 34.2{\scriptsize $\pm$0.5} & 26.2{\scriptsize $\pm$0.5} & 31.4{\scriptsize $\pm$0.7} & 5.2 & 6 \\
\bottomrule
\end{tabular}
}
\end{table}


\begin{table}[htbp]
\centering
\caption{\textbf{Maximum Calibration Error (MCE) across different datasets.} Best scores \textbf{bold}, second-best \underline{underlined}.}
\label{tab:mce_results_reordered}
\resizebox{0.9\textwidth}{!}{
\begin{tabular}{lcccccccc}
\toprule
& \multicolumn{1}{c}{CIFAR10-LT} & \multicolumn{1}{c}{STL10-LT} & \multicolumn{4}{c}{CIFAR100-LT} & Friedman & Final \\
\cmidrule(r){2-2}
\cmidrule(r){3-3}  
\cmidrule(r){4-7}      
\qquad \ \ $\gamma_l$ $\rightarrow$ & 100 & 20 & 20 & 50 & 50 & 50 & Rank & Rank \\
\cmidrule(r){8-9}      
\qquad \ \ $\gamma_u$ $\rightarrow$ & 100 & N/A & 20 & 50 & 1 & 0.02 \\
\qquad \ \ $N_1$ $\rightarrow$ & 500 & 150 & 50 & 150 & 150 & 150 \\
\qquad \ \ $M_1$ $\rightarrow$ & 4000 & N/A & 400 & 300 & 300 & 6 \\
\midrule
FixMatch \cite{sohn2020fixmatch} & 47.5{\scriptsize $\pm$5.0} & 55.1{\scriptsize $\pm$4.9} & 61.3{\scriptsize $\pm$1.8} & 57.3{\scriptsize $\pm$1.1} & 55.3{\scriptsize $\pm$0.8} & 55.5{\scriptsize $\pm$2.4} & 9.0 & 9 \\
+DARP \cite{kim2020darp} & 46.1{\scriptsize $\pm$5.4} & 52.4{\scriptsize $\pm$5.1} & 58.0{\scriptsize $\pm$1.6} & 53.0{\scriptsize $\pm$1.7} & 50.9{\scriptsize $\pm$1.1} & 55.1{\scriptsize $\pm$0.9} & 7.5 &  8 \\
+CReST+ \cite{wei2021crest} & \underline{37.7}{\scriptsize $\pm$5.8} & 48.9{\scriptsize $\pm$5.6} & \textbf{51.0}{\scriptsize $\pm$1.9} & 51.6{\scriptsize $\pm$0.8} & 49.3{\scriptsize $\pm$1.6} & 49.8{\scriptsize $\pm$1.2} &  \underline{3.2} & \underline{2} \\
+ABC \cite{lee2021abc} & 42.1{\scriptsize $\pm$4.6} & 49.2{\scriptsize $\pm$5.1} & 56.4{\scriptsize $\pm$1.2} & \textbf{41.4}{\scriptsize $\pm$1.0} & \underline{40.9}{\scriptsize $\pm$0.4} & \underline{43.1}{\scriptsize $\pm$2.1} & 3.5 & 3 \\
+DebiasPL \cite{wang2022debiasmatch} & 40.0{\scriptsize $\pm$8.8} & 45.2{\scriptsize $\pm$5.1} & 56.0{\scriptsize $\pm$1.6} & 53.7{\scriptsize $\pm$1.1} & 50.4{\scriptsize $\pm$1.1} & 51.9{\scriptsize $\pm$0.8} & 5.2  & 6 \\
+CoSSL \cite{fan2021cossl} & 42.6{\scriptsize $\pm$4.6} & 50.7{\scriptsize $\pm$4.9} & 58.8{\scriptsize $\pm$1.6} & 50.9{\scriptsize $\pm$0.5} & 49.5{\scriptsize $\pm$0.3} & 51.7{\scriptsize $\pm$1.6} & 6.2 & 7 \\
+UDAL \cite{lazarow2023udal} & 41.0{\scriptsize $\pm$5.8} & 50.1{\scriptsize $\pm$5.1} & 56.5{\scriptsize $\pm$1.3} & 51.6{\scriptsize $\pm$0.8} & 49.3{\scriptsize $\pm$1.6} & 49.8{\scriptsize $\pm$1.2} & 4.9 & 5 \\
\rowcolor{Gray}
+ADELLO (ours) & 39.5{\scriptsize $\pm$6.4} & \textbf{25.9}{\scriptsize $\pm$1.0} & \underline{52.8}{\scriptsize $\pm$1.6} & \underline{46.2}{\scriptsize $\pm$0.6} & \textbf{37.9}{\scriptsize $\pm$1.6} & \textbf{42.0}{\scriptsize $\pm$0.5} & \textbf{1.7} & \textbf{1} \\
\midrule
SoftMatch \cite{anonymous2023softmatch} & \textbf{36.8}{\scriptsize $\pm$5.1} & \underline{41.7}{\scriptsize $\pm$6.0} & 56.2{\scriptsize $\pm$2.2} & 53.9{\scriptsize $\pm$1.1} & 45.5{\scriptsize $\pm$2.3} & 50.8{\scriptsize $\pm$1.1} & 3.8 & 4 \\
\bottomrule
\end{tabular}
}
\end{table}

\section{Beyond natural images}
\label{sec:additional_domains}

Following the CIFAR10-LT protocol, we constructed long-tailed versions of TissueMNIST~\cite{medmnistv1}, with 28×28 greyscale \textbf{microscopy medical images} across 8 classes, and EuroSAT~\cite{helber2018eurosat}, featuring 32×32 RGB \textbf{satellite images} in 10 classes.  We use 1/3 of labeled data and all hyper-parameters are set following CIFAR10-LT experiments. Tab.~\ref{tab:sota_new_datasets} shows that our approach can effectively tackle class imbalance and label shift across various image domains.

\begin{table}[htbp]
\vspace{-0pt}
\centering
\caption{Test balanced accuracy (\%) on TissueMNIST-LT and EuroSAT-LT. Comparison of single-classifier approaches.}
\label{tab:sota_new_datasets}
\vspace{-0pt}
\resizebox{0.6\linewidth}{!}{

\begin{tabular}{lcccc}
    \toprule
    & \multicolumn{3}{c}{TissueMNIST-LT} & \multicolumn{1}{c}{EuroSAT-LT} \\
    \cmidrule(r){2-4} \cmidrule(r){5-5}
    $\gamma_l = 100$ / $\gamma_u$ $\rightarrow$ & 100 & $\approx 1$ & 0.01 & 100 \\
    \midrule
    FixMatch \cite{sohn2020fixmatch} & 44.6{\scriptsize $\pm$0.2}       & 45.0{\scriptsize $\pm$0.2}      & 44.7{\scriptsize $\pm$0.3}  & 89.9{\scriptsize $\pm$0.6} \\
    +DARP \cite{kim2020darp} & 44.5{\scriptsize $\pm$0.2}              & 44.5{\scriptsize $\pm$0.1} 
    & 43.9{\scriptsize $\pm$0.5}            & 90.2{\scriptsize $\pm$0.8}    
    \\
    +DebiasPL \cite{wang2022debiasmatch} & 45.2{\scriptsize $\pm$0.5}       & 46.0{\scriptsize $\pm$0.2}      & 45.6{\scriptsize $\pm$0.1}     & 91.8{\scriptsize $\pm$0.4} \\
    +UDAL \cite{lazarow2023udal} & 50.9{\scriptsize $\pm$0.3}       & 51.5{\scriptsize $\pm$0.3}      & 51.4{\scriptsize $\pm$0.1}     & 93.5{\scriptsize $\pm$0.3} \\
    \rowcolor{Gray}
    +ADELLO (ours)     & \textbf{52.3}{\scriptsize $\pm$0.3}       & \textbf{54.3}{\scriptsize $\pm$0.3}      & \textbf{54.4}{\scriptsize $\pm$0.3}     & \textbf{94.1}{\scriptsize $\pm$0.7} \\
    \bottomrule
\end{tabular}

}
\end{table}

\section{Additional algorithmic details}
\label{sec:adello_algorithm}
In Algorithm~\ref{alg:adello_fixmatch_pseudocode}, we provide pseudo-code for ADELLO, utilizing FixMatch as the base SSL algorithm.

\begin{algorithm}
\caption{ADELLO with FixMatch as SSL algorithm}
\label{alg:adello_fixmatch_pseudocode}
\begin{algorithmic}[1]
\State \textbf{Input:} Labeled dataset $D_L$$=$($X_L$, $Y_L$), Unlabeled dataset $D_U$$=$($X_U$, $\cdot$), Model $f$
\State \textbf{Parameters:} Batch size $B$, Batch-ratio $\mu$, Number of classes $K$, Max iterations $t_{\text{total}}$, Confidence threshold $\tau$, Min debiasing factor $\alpha_{\text{min}}$, Schedule speed factor $d$, EMA momentum $\beta$, Warmup iterations $t_{\text{warmup}}$

\Comment{$\sigma$ for softmax$, \omega$ and $\Omega$ for weak and strong data augmentation functions}

\State \textbf{Initialize:} $P_{\text{bal}} \gets (\frac{1}{K},...,\frac{1}{K})$, $\hat{Q} \gets P_{\text{bal}}$, $T \gets 1$

\For{$t = 1$ to $t_{\text{total}}$} \Comment{Main training loop}
    \State $\alpha_t \gets 1.0 - (1.0 - \alpha_{\text{min}}) \cdot \left(\frac{t}{t_{\text{total}}}\right)^d$ \Comment{Update FlexDA target prior}
    \State $\hat{\mathcal{Q}}_{\alpha_t} \gets \text{normalize}(\hat{\mathcal{Q}}^{\alpha_t}$)
    \If{$t = t_{\text{warmup}}$}
        \State $T \gets \text{KL}( P_{\text{bal}}  || \hat{Q} )$ \Comment{Infer temperature $T$ after warmup}
    \EndIf
    
    \State Sample mini-batches $B_L$ from $D_L$ and $B_U$ from $D_U$   
        
    \State $\mathcal{M}(B_u) = \mathbf{1}[\max(\sigma(f(\omega(B_u))), \text{axis}=-1) \ge \tau]$ \Comment{High-confidence mask}
    \State $\mathcal{M}^C(B_u) = 1 - \mathcal{M}(B_u)$ \Comment{Complement mask}
    \State $\hat{y} = \text{argmax}(\sigma(f(\omega(B_u))), \text{axis}=-1)$  \Comment{Predict Hard PLs}
    \State $\tilde{y} = \sigma(\frac{1}{T} f(\omega(B_u)))$  \Comment{Predict Soft PLs}
    
    \State $\mathcal{L}_{s}^{\text{FlexDA}} = \frac{1}{B} \sum^{B}_{b=1} \mathcal{H}(y_b, \sigma(f(\omega(x_b)) + \log \frac{\mathcal{P}_L}{\hat{\mathcal{Q}}_{\alpha_t}} ))$ \Comment{Supervised loss}
    \State $\mathcal{L}_{u}^{\text{FlexDA}} = \frac{1}{\mu B} \sum^{\mu B}_{b=1} \mathcal{M}(u_b) \cdot \mathcal{H}(\hat{y}_b, \sigma( f(\Omega(u_b)) + \log \frac{\hat{Q}}{\hat{\mathcal{Q}}_{\alpha_t}} ))$ \Comment{Consistency loss}
    \State $\mathcal{L}_{uC}^{\text{FlexDA}} = \frac{1}{\mu B} \sum^{\mu B}_{b=1} \mathcal{M}^C(u_b) \cdot \mathcal{H}(\tilde{y}_b, \sigma ( \frac{1}{T} (f(\Omega(u_b)) + \log \frac{\hat{Q}}{\hat{\mathcal{Q}}_{\alpha_t}} )))$ \Comment{CCR loss}
    
    \State $\mathcal{L} = \mathcal{L}_{s}^{\text{FlexDA}} + \mathcal{L}_{u}^{\text{FlexDA}} + \mathbf{1}[t \ge t_{\text{warmup}}] \cdot \mathcal{L}_{uC}^{\text{FlexDA}}$ \Comment{ADELLO objective} 
    
    \State Update $f$ to minimize $\mathcal{L}$

    \State $\hat{Q} \gets \beta \cdot \hat{Q} + (1-\beta) \cdot \text{mean}(\sigma(f(\omega(B_U))), \text{axis}=0)$ \Comment{Update $\hat{Q}$ w/EMA of PLs}
\EndFor
\State \textbf{Output:} Model $f$
\end{algorithmic}
\end{algorithm}

\section{Additional training details}
\label{sec:hyperparameters_settings}
In Table~\ref{tab:hyperparameters}, we provide a comprehensive list of the hyperparameter settings utilized for each dataset. For supervised baselines, the base learning rate starts at 0.1 with a linear warmup. Unless stated otherwise, we reproduce all methods using unified codebases based on \cite{wang2022usb}\footnote{\href{https://github.com/microsoft/Semi-supervised-learning}{https://github.com/microsoft/Semi-supervised-learning} (MIT  license)} for CIFAR10, CIFAR100, and STL10, and based on \cite{fan2021cossl}\footnote{\href{https://github.com/YUE-FAN/CoSSL}{https://github.com/YUE-FAN/CoSSL} (MIT license)} for ImageNet127. 

\begin{table}[htbp]
\centering
\caption{Hyperparameter settings for different datasets.}
\label{tab:hyperparameters}

\resizebox{0.99\textwidth}{!}{

\begin{tabular}{@{}lcccc@{}}
\toprule
Hyperparameter              & CIFAR10-LT                               & CIFAR100-LT                              & STL10-LT                                & ImageNet127                            \\ \midrule
Backbone                       & Wide-ResNet-28-2                         & Wide-ResNet-28-2                         & Wide-ResNet-28-2                        & ResNet-50                               \\
Base SSL algorithm   & FixMatch                                 & FixMatch                                 & FixMatch                                & FixMatch                                \\
Confidence Threshold        & 0.95                                     & 0.95                                     & 0.95                                    & 0.95                                    \\
Optimizer                   & SGD+Nesterov               & SGD+Nesterov                & SGD+Nesterov               & Adam                                    \\
Nesterov Momentum           & 0.9                                      & 0.9                                      & 0.9                                     & -                                     \\
Weight Decay                & 5e-4                                     & 5e-4                                     & 5e-4                                    & -                                     \\
Base Learning Rate          & 0.03    & 0.03     & 0.03    & 0.002                                   \\
Epochs                      & 256                                      & 256                                      & 256                                     & 500                                     \\
Steps per Epoch             & 1024                                     & 1024                                     & 1024                                    & 500                                     \\
Batch Size (labeled)        & 64                                       & 64                                       & 64                                      & 64                                      \\
Batch Size (unlabeled)      & 128                                      & 128                                      & 128                                     & 64x2 views                                    \\
FlexDA \(\alpha_{\text{min}}\)     & 0.1                                      & 0.1                                      & 0.1                                     & 0.1                                     \\
FlexDA \(d\)                       & 2                                        & 2                                        & 2                                       & 2                                       \\
FlexDA EMA $\beta$ & 0.999 & 0.999 & 0.999 & 0.999 \\
Temperature $T$ & inferred & inferred & inferred & inferred \\
Warm-up $t_{\text{warmup}}$            & 50k                           & 50k                           & 0                                    & 0                                    \\
\(\lambda_u\) & 1                                        & 1                                        & 1                                       & 1                                       \\ 
\(\lambda_{uC}\)& 1                                        & 1                                        & 1                                       & 1                                       \\ 
\bottomrule
\end{tabular}
}

\end{table}

\end{document}